%% file: main.tex
\documentclass[prologue,table,xcdraw,sigconf]{acmart}
\AtBeginDocument{%
  \providecommand\BibTeX{{%
    \normalfont B\kern-0.5em{\scshape i\kern-0.25em b}\kern-0.8em\TeX}}}

\copyrightyear{2022} \acmYear{2022}
\setcopyright{rightsretained}
\acmConference[MM '22]{Proceedings of the 30th ACM International Conference on Multimedia}{October 10--14, 2022}{Lisboa, Portugal} \acmBooktitle{Proceedings of the 30th ACM International Conference on Multimedia (MM '22), October 10--14, 2022, Lisboa, Portugal}\acmDOI{10.1145/3503161.3548287} \acmISBN{978-1-4503-9203-7/22/10}

\input{preamble}
\begin{document}
\title{Action-conditioned On-demand Motion Generation}

\author{Qiujing Lu}
\affiliation{
   \institution{University of California, Los Angeles}
   \country{USA}
}
\email{qiujing@g.ucla.edu}
 
\authornote{These authors contributed equally to this work} 

\author{Yipeng Zhang} \authornotemark[1]
\affiliation{
   \institution{University of California, Los Angeles}
   \country{USA}
}
\email{zyp5511@g.ucla.edu}

\author{Mingjian Lu}
\affiliation{
   \institution{University of California, Los Angeles}
   \country{USA}
 }
\email{mingjianlu@g.ucla.edu}

\author{Vwani Roychowdhury}
\affiliation{
   \institution{University of California, Los Angeles}
   \country{USA}
}
\email{vwani@g.ucla.edu}


\input{sec/0_abstract}




\input{figure/teaser}
\maketitle
\input{sec/1_introduction}
\input{figure/our_arch}
\input{sec/2_related}
\vspace{-0.5cm}
\input{sec/3_method}

\input{sec/4_results}
\input{sec/5_conclusions}

\newpage
{
    \small
    \bibliographystyle{ACM-Reference-Format}
    \bibliography{macros,main}
}

\input{sec/supplimentary}


\end{document}

%% file: preamble.tex
\usepackage{overpic}

\usepackage{enumitem} 
\usepackage{overpic} 
\usepackage{color}
\usepackage{multicol}
\usepackage[export]{adjustbox}

\definecolor{turquoise}{cmyk}{0.65,0,0.1,0.3}
\definecolor{purple}{rgb}{0.65,0,0.65}
\definecolor{dark_green}{rgb}{0, 0.5, 0}
\definecolor{orange}{rgb}{0.8, 0.6, 0.2}
\definecolor{red}{rgb}{0.8, 0.2, 0.2}
\definecolor{darkred}{rgb}{0.6, 0.1, 0.05}
\definecolor{blueish}{rgb}{0.0, 0.3, .6}
\definecolor{light_gray}{rgb}{0.7, 0.7, .7}
\definecolor{pink}{rgb}{1, 0, 1}
\definecolor{greyblue}{rgb}{0.25, 0.25, 1}
\usepackage{ulem}
\usepackage{booktabs}

\usepackage{blindtext}

\renewcommand{\paragraph}[1]{\vspace{1em}\noindent\textbf{#1}.}

\def\1{\mathbbm{1}}
\DeclareMathOperator{\Tr}{Tr}
\usepackage{amsfonts, cleveref}
\usepackage{bm}
\usepackage{bbm}
\newcommand{\myvec}[1]{\bm{\mathrm{#1}}}
\newcommand{\mstd}[2]{{#1}^{\pm{#2}}}
\newcommand{\norm}[1]{\left\lVert#1\right\rVert}

\usepackage{wrapfig}

\useunder{\uline}{\ul}{}


%% file: sec/0_abstract.tex
\begin{abstract}
We propose a novel framework, On-Demand MOtion Generation (ODMO), for generating realistic and diverse long-term 3D human motion sequences conditioned only on action types with an additional capability of customization. ODMO shows improvements over SOTA approaches on all traditional motion evaluation metrics when evaluated on three public datasets (HumanAct12, UESTC, and MoCap). Furthermore, we provide both qualitative evaluations and quantitative metrics demonstrating several first-known customization capabilities afforded by our framework, including mode discovery,  interpolation, and trajectory customization.These capabilities significantly widen the spectrum of potential applications of such motion generation models. The novel on-demand generative capabilities are enabled by innovations in both the encoder and decoder architectures: (i) Encoder: Utilizing contrastive learning in low-dimensional latent space to create a hierarchical embedding of motion sequences, where not only the codes of different action types form different groups, but within an action type, codes of similar inherent patterns (motion styles) cluster together, making them readily discoverable; (ii) Decoder: Using a hierarchical decoding strategy where the motion trajectory is  reconstructed first and then used to reconstruct the whole motion sequence. Such an architecture enables effective trajectory control. Our code is released on the Github page: \href{https://github.com/roychowdhuryresearch/ODMO}{\textcolor{blue}{https://github.com/roychowdhuryresearch/ODMO}}
\end{abstract}

%% file: figure/teaser.tex
\begin{teaserfigure}
\begin{center}
\includegraphics[width=0.74\linewidth]{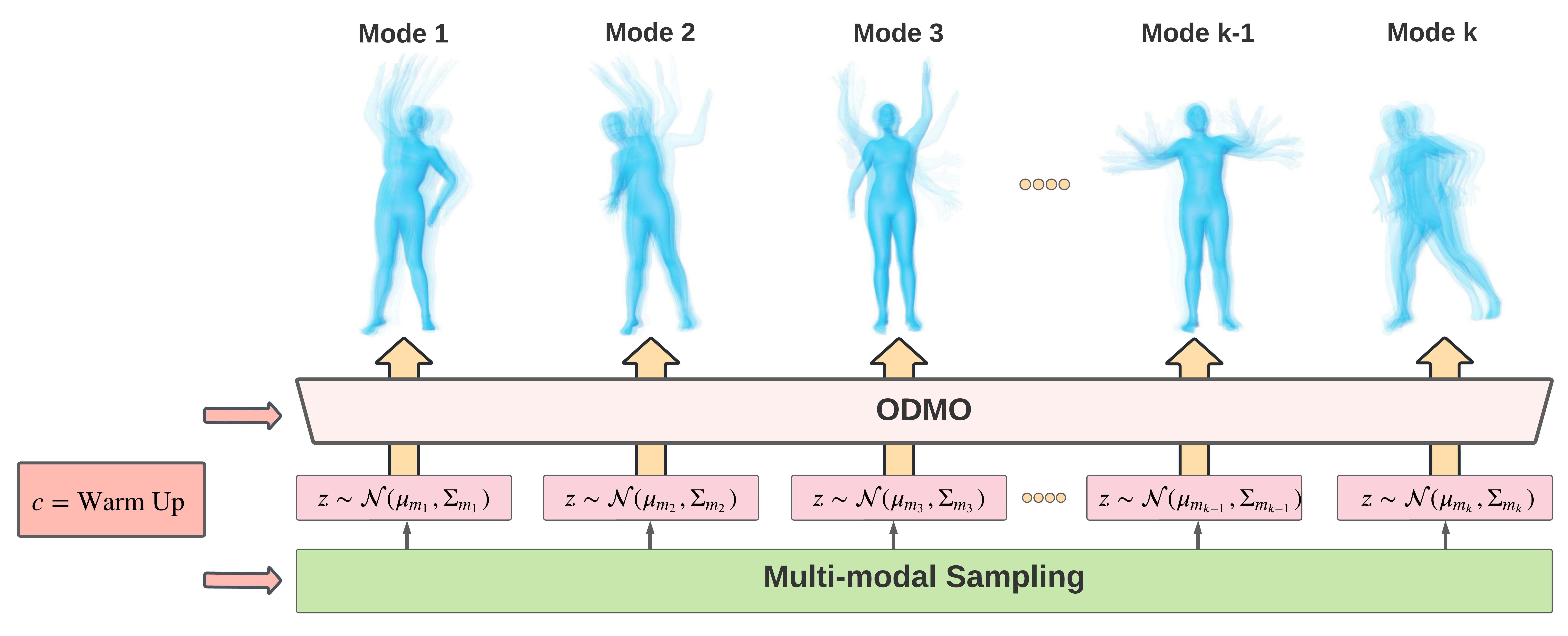}
\end{center}
\caption{On-Demand MOtion Generation(ODMO) can generate diverse and realistic 3D human motion sequences conditioned on action type $c$ with the capability of automated discovery and controllable generation of different subtypes in a given category. 
}
\label{fig:teaser}
\end{teaserfigure}

%% file: sec/1_introduction.tex
\section{Introduction}
\label{sec:intro}
A much coveted goal in multimedia studies is the design of a generator software that can produce animated 3D human characters --on demand, executing specific activities-- which can be seamlessly integrated into an existing multimedia document. If such a holy grail can indeed be achieved, then it would revolutionize the field of  automated generation  of multimedia content, enabling an average user to easily express visual memories, scenes, or dreams, and express their creativity. While we are far away from such an ideal multimedia oracle, the advent of deep learning has seen several impressive attempts at automating smaller steps towards this goal. For example, translations of different modalities, such as text, speech, music, to 2D motion sequences have been attempted. Due to the difficulties of collecting data, most of them are conducted in 2D space or can only generate motions that are short in duration. 

While considerable progress has been made in pixel-based video generation \cite{vondrick2016generating, huang2018human, xu2021progressive}, the resulting human appearance is still unsatisfactory \cite{guo2020action2motion}. A fundamental step to reduce such  distortions is to use 3D human skeleton sequences to guide  video generation. A wide variety of frameworks with different  assumptions about prior information have been developed for the resulting 3D motion sequence generation task. For example, in the motion prediction  task\cite{aliakbarian2021contextually, cai2020learning, tang2018long, 9157765, Zhang:CVPR:2021, Cui_2021_CVPR, yuan2020dlow}, initial frames with optional control signals are provided as inputs; similarly, in a motion transition generation task, several temporally-sparse keyframes are used as inputs \cite{harvey2020robust}; finally, in a motion style transfer and synthesis task,  
 a pair of motion sequences are used as inputs \cite{wen2021autoregressive}. Despite their promising performance, initial or intermediate sub-sequences are still required as inputs, which limit these models' applications.

We address the fundamental problem of 3D motion generation without any initial motion as inputs. Given only the activity type, chosen from among a pool of different activities, and its characteristic features, the generative model should generate motion sequences that are physically realistic and diverse in style, reflecting variations across the human population. Amongst many related works such as \cite{pavllo2018quaternet, wang2021synthesizing, xia2019learning}, the most relevant and recent ones are two stage GAN \cite{cai2018deep}, MoCoGAN \cite{tulyakov2018mocogan}, Action2Motion \cite{guo2020action2motion} and ACTOR \cite{petrovich21actor}.
\cite{guo2020action2motion} in particular is a striking recent work 
that is among the first to propose automated generation of 3D pose sequences for multiple activities from the same generator, including complex motion such as dancing and animal behavior. 
A recent follow up work \cite{petrovich21actor}, applies a Transformer-based variational autoencoder (VAE) architecture (instead of being LSTM based) for encoding and decoding, and avoids a time-dependent latent space embedding as done in \cite{guo2020action2motion}; it shows considerable performance gain over \cite{guo2020action2motion}. In both these frameworks \textit{addressing our problem of motion generation based only on action type}, customization capabilities such as the ability to generate different styles on demand (i.e. in a controllable fashion) within the same category or customization of the motion trajectory via the destination are missing. 
 
Our work has the following distinctive design features:

\textbf{1) Internal causal motion representation}: Any 3D sample motion is decomposed into two parts: spatial joint movements relative to a body center (root) -- referred to as local movement profile (LMP)-- and the resulting trajectory of the body center. 
This key step is rooted in the kinematics of body movement: We move our body parts relative to each other (as captured in the LMP), and these movements after interaction with the environment cause our body center to have spatial movement resulting in a particular trajectory. We, thus, first encode the primary causal factor (the LMP) into a low-dimensional latent code, and then the latent code is decoded in a hierarchical fashion.

\textbf{2) Hierarchical decoding:} In the decoding procedure, a matching trajectory is first reconstructed from the latent code, enforcing the trajectory generator to learn how the environment interacts with the  movements of body parts. Then, the entire motion sequence is generated by integrating the latent code and the reconstructed trajectory. The hierarchical decoding enables us to control the trajectory generation necessary for  accurate trajectory customization.  We believe that this pipeline also enables us to perform competitively even with the $(x,y,z)$ representations of the skeleton joints, allowing our model to work with rich datasets where only 3-D representations are available or reliably computable. 

\textbf{3) Multi-modal latent code space engineering}: Additional regularization terms are added to the VAE-type encoder loss function to enforce a hierarchical latent space, where codes belonging to different action types are separated into distinct clusters and within each such cluster the similar modes and styles present in real motion group together into more compact clusters. This is achieved via contrastive learning during training.

The design choices enable several customization capabilities including the following:
\textbf{1) Multiple mode discovery and multi-modal sampling}: Since the codes for action categories are separated, we do a Gaussian Mixture Model (GMM) fitting of the codes in the same category, where the number of mixture components is determined adaptively, to discover different styles within an action type. These distributions are then sampled to generate diverse motions on demand. \textbf{2) Intra-action type motion interpolation}: Given the multi-modal representations of the latent code space for each action type, one can perform interpolation between modes to generate realistic mixed motion sequences not seen in the training set. \textbf{3) Trajectory customization via destination location (end-point) specification:} Realistic and controllable motion sequences can be generated by varying the latent code and the destination location. By varying the destination location, while keeping the same latent code (hence, style) motion sequences that cover a large spatial domain are generated; analogously, our framework enables one to generate motion with different styles (by varying the code) while keeping end-point the same; see \Cref{fig:multim}. At the population level, such generated motion can be shown to be both realistic and customizable by benchmarking motion quality metrics.

%% file: figure/our_arch.tex
\begin{figure*}[!h]
\centering
    \includegraphics[width = 0.95\textwidth]{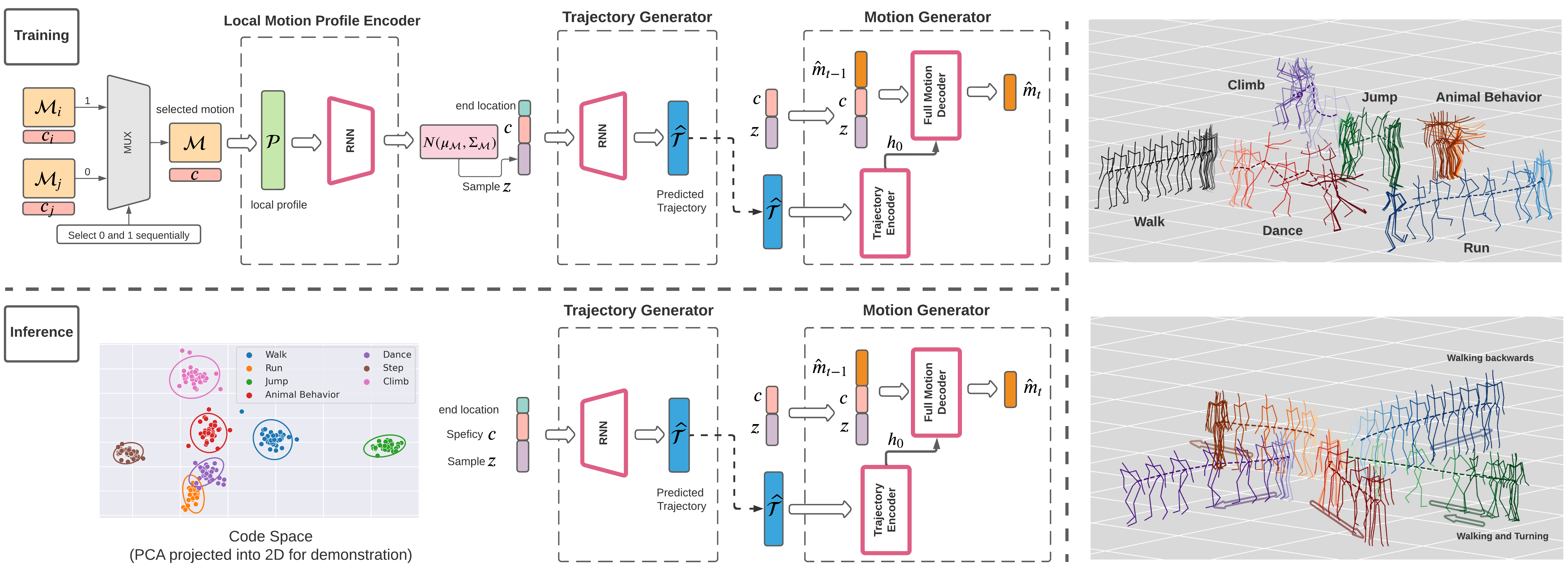}
    \caption{(Left) Overview of ODMO model architecture. The loss function for the pair $\{\mathcal{M}_i, \mathcal{M}_j\}$  has a contrastive term, which encourages different classes and styles to form distinct clusters in the latent space. The illustrated code space is generated from randomly sampled $10\%$ of real motion from MoCap dataset. For visualization,  we fitted a Gaussian distribution per category and adopted the dimension reduction (PCA) to project the latent codes into 2D, with their associate labels color-coded. The ellipse around each cluster is the contour line of the accordingly projected Gaussian distribution at half-peak value. (Right) Motion sequences generated for multiple activities and diverse Walk activity motion sequences in MoCap dataset.}
    \label{fig:arch}
\end{figure*}

%% file: sec/2_related.tex
\section{Related work}
\label{sec:related}

\textbf{Modeling human motion:}
As already mentioned, various methods have been proposed to model and generate motion with widely varying goals and assumptions about prior information. The more traditional models, such as Hidden Markov Models \cite{tanco2000realistic}, switching linear dynamic systems \cite{pavlovic2000learning}, Gaussian processes \cite{wang2007gaussian} and heterogeneous features \cite{7784788} have been proposed. However, some of these methods, like Gaussian processes, are well-known to be computationally intractable. More recently, deep learning frameworks have also been used for action modeling \cite{9706375,Li_2020_CVPR} and deterministic or probabilistic motion generation: deep recurrent neural networks (RNNs)  for modeling dynamic temporal behavior  \cite{zhou2018auto, fragkiadaki2015recurrent, martinez2017human, wang2019combining}; Autoencoder \cite{holden2015learning}, Normalising Flows \cite{henter2020moglow}, VAE \cite{yan2018mt} and GAN \cite{tulyakov2018mocogan}. Additionally, embedding of motion in a low dimensional latent space is used implicitly or explicitly by several models to assist their generation tasks, including both time dependent \cite{wang2007gaussian,yan2019convolutional,guo2020action2motion, hassan2021stochastic,rempe2021humor} or independent \cite{holden2015learning,petrovich21actor} encodings. 

A frame in a given motion sequence is often encoded, especially in rotation based representations, as local motion with respect to a root node (usually the pelvis) and the 3D coordinate of the pelvis \cite{petrovich21actor,Villegas_2018_CVPR,villegas2021contact}. The input to the model, however, comprises both these components. In our work, we encode only the LMP of the motion sequences, without the trajectory part.  The trajectory is then decoded from the corresponding LMP code, and final motion sequence is generated jointly by the predicted trajectory and the code. Similar motion decomposition has been successfully used in other domains, such as motion inbetweening \cite{zhou2020generative}, but it is the first time applying this scheme effectively in our domain of action-conditioned motion generation task. 

\textbf{Human motion prediction:} Similar to the motion generation, short and long term motion prediction also require modeling of human motion. However, they require additional information as input: the first several frames of the desired activity. Moreover, the evaluation metrics are mainly designed to compare the predicted future motion with ground truth motion. In our setup, the generated motion sequences can only be evaluated at the population level with the ground truth data. \textit{Deterministic prediction:}
\cite{9157765, Cui_2021_CVPR} both use Residual Graph Convolutional Network (GCN) to capture spatial-temporal correlation and rely on history information to generate \textit{deterministic} single future motion. \textit{Stochastic prediction:}
\cite{yuan2020dlow} introduces a sampling version of motion prediction, and also considers diversity in the forecast motion. They use learnable mapping functions to map samples from a single random variable (generated by a given historical information sequence) to a set of correlated latent codes for further decoding into a set of correlated future motion sequences. \cite{Zhang:CVPR:2021} uses novel VAE which preserves the full temporal resolution of the history motion as condition and samples from the latent frequencies explicitly to introduce high-frequency components into the generated motion. \cite{aliakbarian2020stochastic,aliakbarian2021contextually} are both based on Conditional VAE with novel mechanisms to enforce diversity and contextual consistency with the history information for the final prediction. However, in such stochastic prediction tasks, the past motion plays an important role (as input and condition). 
To illustrate this, \textit{we modify the motion prediction model in \cite{yuan2020dlow} to suit our generation task and set it as a baseline algorithm.}\\

%% file: sec/3_method.tex
\section{Method}
\subsection{Notations}
 Given a set of 3D motion sequences, $\{ \mathcal{M}_1, \mathcal{M}_2, \ldots, \}$ along with their corresponding activity categories $c \in C$ ($|C|$ is the number of categories), our goal is to train a generative model that is able to produce 3D motion sequences with fixed length $T$ for a specified activity category. Inside a motion sequence $\myvec{\mathcal{M}} = [\myvec{m}_1, \myvec{m}_2 ,\ldots , \myvec{m}_T]$, the pose $\myvec{m}_t$ in the $t^{th}$ frame  is represented by the locations (e.g. their $(x,y,z)$ coordinates) of $J$ joints:  $\myvec{m}_t = [\myvec{m}_t(0),\ldots, \myvec{m}_t(J-1)]^T \in \mathbb{R}^{3 \times J}$. One joint, usually the center of the pelvis is referred to as the root joint.  We assume that the body center can be approximated by the root joint, and its movement in 3D space along time will be referred as the \textit{trajectory of the motion}, and  denoted as $\mathcal{T} = [\myvec{r}_1, \ldots, \myvec{r}_T ]$ , where $\myvec{r}_t \in \mathbb{R}^3$.  The end location $\myvec{r}_T$ of the root joint will be referred to as $\myvec{r_e} \in \mathbb{R}^3$, and can be used to guide motion generation in our model.  Additionally, each activity class $c$ is denoted by a one-hot encoded vector. Furthermore, \textit{local movement profile} (LMP) of a given motion sequence $\mathcal{M}$ is defined as the locations of the joints relative to the root joint position, i.e. $\mathcal{P}= [\myvec{p}_1, \myvec{p}_2,\ldots, \myvec{p}_T]$, where $\myvec{p}_t \in \mathbb{R}^{3 \times J}$, and $\myvec{p}_t(i)= \myvec{m}_t(i) - \myvec{r}_t,\,i=0,\ldots, J-1,\,t= 1, \ldots, T$. This decomposition of any motion sequence $\mathcal{M}$ into two components, the LMP $\mathcal{P}$ and the resulting trajectory $\mathcal{T}$, enables us to establish a causal relationship between body movement of the human agent, its environment, and the trajectory that ensues. 

\subsection{Model Architecture and Loss Functions}
There are three main components in our ODMO (\textbf{O}n \textbf{D}emand \textbf{MO}tion generator): local movement profile encoder (LMP encoder), trajectory generator, and motion generator.
\footnote{For simplicity, the loss functions introduced in this section are defined for a single sampled motion; the only exception is $\mathcal{L}_{es}$ defined in Equation~\ref{eq:contrastive}, which involves a pair of sampled motions, $\mathcal{M}_1$ and $\mathcal{M}_2$, whose corresponding LMPs are $\mathcal{P}_1$ and $\mathcal{P}_2$, and are of class $c_1$ and $c_2$}:

i) LMP encoder $\mathrm{Es}_{
\theta}$, parametrized by network parameters $\theta$ which learns a representation of local movement profiles ($\mathcal{P}$) in a lower dimensional space $\mathbb{R}^{D}$. Samples for training the decoder, $\mathbf{z}$, are drawn from $\mathbf{z} \sim \mathcal{N}\left(\mu_{\theta}, \Sigma_{\theta}\right)$, where  $\left(\mu_{\theta}(\myvec{\mathcal{P}}), \Sigma_{\theta}(\myvec{\mathcal{P}})\right)=\mathrm{Es}_{\theta}(\myvec{\mathcal{P}})$. A contrastive loss is applied to a pair of LMPs, $(\myvec{\mathcal{P}}_1, \myvec{\mathcal{P}}_2)$, according to Equation \ref{eq:contrastive}. The goal of the Equation \ref{eq:contrastive} is to create a multi-cluster distribution in
the latent space, where not only the embeddings of motion sequences belonging to different categories form distinct clusters, but also the codes of hidden modes within each such cluster are separated into smaller clusters. These clusters can be modeled as GMMs, which are then sampled to generate motion (see \Cref{sec:mm_sampling}). This joint goal is achieved by (i) contrastive margin $\alpha$ that separates the action types, and (ii) the hyper-parameter $\sigma^2$ from the prior distribution of $\Sigma_{\theta}$ that preserves the multimodality inherent in real motion (see \textit{Supplementary Materials} for detailed derivation). 

\begin{align}
\label{eq:contrastive}
\mathcal{L}_{es}(\theta) &= \1_{\{c_1=c_2\}} \left\| \mu_{\theta}(\myvec{\mathcal{P}}_1)-  \mu_{\theta}(\myvec{\mathcal{P}}_2) \right\|_{2}^2 \nonumber\\
&+ \1_{\{c_1\neq c_2\}}\; {\max\left\{0,\; \alpha -\left\|\mu_{\theta}({\myvec{\mathcal{P}}_1}) -  \mu_{\theta}(\myvec{\mathcal{P}}_2)\right\|_{2}\right\}}^2\nonumber\\
& - \ln(\det \Sigma_{\theta}(\myvec{\mathcal{P}}_1)) + \frac{1}{\sigma^2}\Tr(\Sigma_{\theta}(\myvec{\mathcal{P}}_1))\nonumber\\ &- \ln(\det \Sigma_{\theta}(\myvec{\mathcal{P}}_2)) + \frac{1}{\sigma^2}\Tr(\Sigma_{\theta}(\myvec{\mathcal{P}}_2))
\end{align}

ii) Trajectory generator $\mathrm{Gr}_{\phi}$, which predicts the resulting 3D trajectory $\mathcal{T}$ via predicting trajectory velocities
for $T$ steps
, based on the sampled style code $\mathbf{z}$ and one-hot encoded categorical information $c$. The predicted trajectory $\mathcal{\hat{T}}$ will be reconstructed based on the accumulated velocity predictions, i.e. $\hat{\myvec{r}}_{t} =\left(\mathrm{Gr}_{\phi}\left(\mathrm{c},\mathbf{z}, \myvec{r_e} \right)\right)_t + \hat{\myvec{r}}_{t-1}$, where $\hat{\myvec{r}}_t$ is the predicted location at time $t$ ($\hat{\myvec{r}}_0=\myvec{0}$). The trajectory generator is trained to predict the velocity of root trajectory $\myvec{r}_t -  \myvec{r}_{t-1}$ at $t^{th}$ time step ($\myvec{r}_0=\myvec{0}$) with loss: 
\begin{align}
\label{eq:traj_gen}
\mathcal{L}_r(\phi) &= \sum_{t=1}^T\left\| \left(\mathrm{Gr}_{\phi}\left(\mathrm{c},\mathbf{z}, \myvec{r_e} \right)\right)_t - (\myvec{r}_t -  \myvec{r}_{t-1})\right\|_{2}^2.
\end{align}
Note that in this formulation we use end location information as an input to enable trajectory customization as described in \Cref{subsec:traj_cus}. However, we also train models without any end location as $-\myvec{r_e}$ in \Cref{section:ablation}. As shown there, the standard ODMO model preserves performance while providing customization capabilities. 

(iii) Motion generator $\mathrm{Gm}_{\psi}$, which generates realistic motion sequence $\hat{\mathcal{M}}$ based on style code $\mathbf{z}$, guidance trajectory $\hat{\mathcal{T}}$, category label $c$. The motion generator is adapted from Seq2Seq \cite{seq2seq} which consists of an encoder module and decoder module. The trajectory encoder encodes the predicted trajectory and sends the compact representation of the trajectory to the full motion decoder (decoder) as the initial hidden state  $h_0$.  The decoder predicts the whole motion sequence $\hat{\myvec{m}}_1, \hat{\myvec{m}}_2 ,\ldots , \hat{\myvec{m}}_T$ sequentially, as in Equation \ref{eq: gm}. 
At $t^{th}$ step of decoding, the decoder employs the code $\mathbf{z}$, class information $c$, and the previous step of output $\hat{\myvec{m}}_{t-1}$ (or $\myvec{m}_{t-1}$ if teacher forcing is used) to predict the pose $\hat{\myvec{m}}_{t}$.  In the training process, we apply the reconstruction loss $\mathcal{L}_{mre}$ on the predicted motion, 
\begin{align}
\label{eq: gm}
\mathcal{L}_{mre}(\psi) &= \sum_{t=1}^T \left\|\myvec{m}_t - \mathrm{Gm}_{\psi}\left(\tilde{\myvec{m}}_{t-1}, c,\myvec{z}, \hat{\myvec{r}} \right) \right\|_{2}^2, \nonumber\\
&\tilde{\myvec{m}}_{t-1} = \begin{cases}
\myvec{m}_{t-1}, \text{ if tf}=1 \\
\hat{\myvec{m}}_{t-1}, \text{ otherwise}.
\end{cases}
\end{align} 
A consistency loss is also applied to ensure predicted trajectories from trajectory generator and motion generator are coordinated:
\begin{align}
\label{eq:cons}
\mathcal{L}_{cons} = \sum_{t=1}^T \left\|\hat{\myvec{r}}_{t} - \hat{\myvec{m}}_{rt} \right\|_2^2,
\end{align} where we denote the predicted trajectory from the motion generator $\mathrm{Gm}_{\psi}$ as $\hat{\myvec{m}}_{r1},\ldots , \hat{\myvec{m}}_{rT}$. As ablation studies in \Cref{subsec:traj_cus} show, this term is essential for accurate trajectory customization. 

Finally, we combine the loss terms in \Cref{eq:contrastive,eq:traj_gen,eq: gm,eq:cons} to jointly optimize the whole network: 
\begin{align}
\label{func: final loss}
\mathcal{L}(\theta, \psi, \phi) & =\mathcal{L}_{es}(\theta) + \mathcal{L}_r(\phi) + \mathcal{L}_{mre}(\psi) +
\mathcal{L}_{cons}
\end{align}.

\textbf{The general workflow of ODMO}: As shown in \Cref{fig:arch}, in the training process, at each time step, a pair of randomly picked motion sequences $\{\mathcal{M}_i, \mathcal{M}_j\}$ are sent into the model sequentially.  The processing of each $\mathcal{M}$ is like in a VAE, except that since the local movement profile is encoded by the encoder, our decoding step requires both trajectory and motion generators to reconstruct the final motion. During the inference process, only the desired class $c$ is specified. Given $c$, the code space is sampled to generate a code $z$, and a corresponding end-point $\myvec{r_e}$. Then these are fed to the rest of the pipeline, where the full motion decoding is in an auto-regressive manner. 

\subsection{Sampling the Latent Code Space}
\label{sec:mm_sampling}
\textbf{Mode-preserving sampling}: Our contrastive learning framework enables one to flexibly organize the latent codes into different sets of distributions and result in a variety of sampling strategies suited for different applications.  
In real motion, there would be several modalities in each motion category; for example, in the Jump category, one can have humanoids jumping forward, or jumping vertically with little horizontal motion. 
Our training formalism encodes the modalities of an action category (present in the data) as computationally discoverable modes in the distribution of latent codes inferred for that category. 
We fit a GMM with $K$ components to the codes for each category, where $K$ is chosen adaptively by using the silhouette score \cite{ROUSSEEUW198753}. One can sample these action specific GMMs to generate multi-modal motion for the same action type. Such GMM  allows us to generate samples from modes proportionally to the number of motion in each discovered mode for each category. We call this mode-preserving sampling strategy. We, however, can tailor the probabilities of the modes (choosing them differently from the discovered mode probabilities in the training set) in the GMM to create new distributions of sampled data. This can find additional applications in data augmentation of downstream tasks \cite{petrovich21actor}. 

The conventional sampling process based solely on category information and samples from $\mathcal{N}(\mathbf{0}, \mathbf{I})$, which is not mode aware, can also be replicated by directly fitting $|C|$ GMMs to all the codes of the real motion data. Each such Gaussian can then be sampled to generate motion sequences.  We demonstrate the results of such sampling as ODMO$_{conv}$ in the \Cref{tab:baseline_benchmark,tab:baseline_mocap}; its performance is comparable to ODMO, but it lacks understanding of modes and mode-aware customization ability.

%% file: sec/4_results.tex
\input{tab/mingjian_trial}
\input{tab/mocap}
\section{Experiment}
Our base model, ODMO adopts the mode-preserving GMM sampling strategy. The end location is generated by a k-nearest neighbors (KNN) trained with codes from real motion and their corresponding end locations.

\subsection{Datasets}

We evaluated models on three public motion datasets:

\textbf{HumanAct12:}
We used the same HumanAct12 dataset as introduced in \cite{guo2020action2motion, zou2020polarization}
which consists of 1,191 motion clips and 12 categories such as: Warm up, Walk and Drink. 

\textbf{UESTC:} Compared to HumanAct12 dataset, UESTC \cite{ji2018large}  uses a finer granularity in defining action categories, resulting in 40 action types performed by 118 subjects; example action types include:  Squatting, Punching and Left stretching. There are 25,600 video samples collected from 8 fixed viewpoints. For fair comparison, we used the same preprocessed data as \cite{petrovich21actor}. Motion sequence samples shorter than a given threshold are filtered. This results in 10,629 training sequences and 13,350 testing sequences. 

\textbf{CMU MoCap (MoCap):}
We identified seven action categories such as Walk, Run and Dance from the CMU MoCap data library consisting of 970 motion sequences in total. For fair comparison, we used the same data preprocessing steps as \cite{guo2020action2motion} but dropped the Wash activity, as most of the motion sequences inside this class are mislabeled.

Among these three datasets, MoCap has real motion with longer overall spatial translations as well as more diversity within each action type making it much more challenging.
\subsection{Results}
 
\subsubsection{Metrics based on classifiers}\label{subsubsec:metrics-classif}
In this approach, an action-category classifier is designed based on real motion, and then is used to evaluate different ensemble properties of the generated motion. The classifier metrics for the real motion then can be used as ground truth. Adopted from \cite{petrovich21actor,guo2020action2motion, lee2019dancing}, the metrics we used are as follows:
Recognition Accuracy (\textbf{Acc}) measures the confidence of generated motion belonging to a given category.  Higher Acc implies better motion quality. Frechet Inception Distance (\textbf{FID}) \cite{heusel2017gans} computes the distance between the distributions of the classifier features of the real motion and the generated motion; hence, \textit{lower FID is desirable}. The  \textbf{Multimodality} (\textbf{Diversity}) is the expected distance  between classifier features of two randomly sampled generated motion sequences belonging to the same action category (across all categories). Thus,   \textit{motion samples generated by a better model should have   multimodality/diversity closer to that of the real motion}.  

We trained such classifiers with real motion sequences as mentioned in \cite{guo2020action2motion, petrovich21actor}. For MoCap, we re-trained the classifier with the same architecture as in \cite{guo2020action2motion} but changed the number of outputs to $7$. For HumanAct12, we directly employ the same classifier as \cite{guo2020action2motion, petrovich21actor}. For UESTC, since the classifier from \cite{petrovich21actor} is in $6$D representation, it is not suitable for our xyz representation. Hence, we followed the train and test split as in \cite{petrovich21actor} and trained our own classifier in xyz representation (the architecture is shown in the \textit{Supplementary Materials}). The metrics obtained from the classifiers are reported in \Cref{tab:baseline_benchmark,tab:baseline_mocap} as the mean of $10$ sets of samples generated from each model with the $95\%$ confidence interval on the top; each set comprises $400/250/200$ samples for each action type in Mocap/HumanAct12/UESTC dataset.
\subsubsection{Alternative diversity metric} 
Besides the multi-modality metric used in the classifiers, there is another popular metric to measure diversity in any given set $\mathcal{A}= \{\myvec{m}^{(1)}, \dots, \myvec{m}^{(N)}\}$ of $N$ motion sequences introduced in \cite{yuan2020dlow}: the average pairwise distance (APD). It can be computed as follows:
$$
\mathrm{APD}(\mathcal{A})=\displaystyle \frac{1}{N(N-1)}{\sum_{i=1}^N\sum_{j\neq i}^N(\sum_{t=1}^T\|\myvec{m}_t^{(i)}-\myvec{m}_t^{(j)}\|^2)^{\frac{1}{2}}}.$$ 
Given two sets of motion sequences $\mathcal{A}$ and $\mathcal{B}$, if APD($\mathcal{A}$) $<$APD($\mathcal{B}$) then $\mathcal{A}$ represents a more homogeneous set of motion sequences than $\mathcal{B}$; equivalently, $\mathcal{B}$ can be said to represent a more diverse set of motion sequences.
We adapted this APD metric to investigate diversity across all action categories. 
For each model, $\mathcal{I}$, and for each category $C_i$ in a dataset, let $\mathcal{M}_{\mathcal{I}} (C_i)$ be the set of generated motion sequences. We first compute APD($\mathcal{M}_{\mathcal{I}} (C_i)$). Since there is a high variance in this APD metric based on the action type $C_i$, we also create a motion sample set from the ground truth motion sequences: $\mathcal{M}_{\mathcal{G}} (C_i)$. Then we define a normalized APD, n-APD($\mathcal{M}_{\mathcal{I}} (C_i)$), as the ratio of APD($\mathcal{M}_{\mathcal{I}} (C_i)$) and APD($\mathcal{M}_{\mathcal{G}} (C_i)$). Finally, the n-APD metric for a model $\mathcal{I}$, n-APD($ \mathcal{I}$), is defined as the average of n-APD($\mathcal{M}_{\mathcal{I}} (C_i)$) over all categories $C_i$.
$$ \text{n-APD(}\mathcal{I}\text{)} =\sum_{i=1}^{|C|} \frac{1}{|C|}\frac{APD(\mathcal{M}_I(C_i))}{APD(\mathcal{M}_{\mathcal{G}}(C_i))}$$
We reported  n-APD($\mathcal{I}$) (n-APD) for all models($\mathcal{I}$) using the same data generated as in \Cref{tab:baseline_benchmark,tab:baseline_mocap}.  
\input{figure/demo_mm_fig}
\subsubsection{Comparison with SOTA models}

Action2Motion \cite{guo2020action2motion} (including xyz (A2M-xyz) and Lie group representation (A2M-Lie)) and ACTOR \cite{petrovich21actor} are the only two prior works for action-conditioned motion generation task. However, both methods use the VAE framework for motion generation, so we also include a GAN model as modified in \cite{guo2020action2motion}, MoCoGAN\cite{tulyakov2018mocogan} (Act-MoCoGAN). Since \cite{guo2020action2motion} does not provide their trained Act-MoCoGAN model, we trained it on our three datasets following their architecture.

Additionally, since probabilistic prediction and generation tasks both generate diverse motion sequences, we also consider a slightly modified version of a SOTA probabilistic prediction model, DLow\cite{yuan2020dlow} (DLow-Gen): we replaced the initial frames of the given motion sequence with its activity category label.

To compare performance on the three datasets, we did the following: (i) ACTOR \cite{petrovich21actor}: For both HumanAct12 and UESTC, the sample motion sequences were generated using their released models. As already mentioned, for uniform comparison across all models all the classifiers are trained using xyz representation, and hence ACTOR's outputs in $6$D representation were converted to xyz, so that all the baselines can be evaluated using the same classifiers.  
This conversion was done following the same implementation used in \cite{petrovich21actor}. For the MoCap dataset provided by \cite{guo2020action2motion}, the motion representation is in xyz. On training the ACTOR model on xyz, or on converted $6$D, it did not yield competitive performance. (ii) Action2Motion \cite{guo2020action2motion}: For HumanAct12, we directly used their released models;  For UESTC and MoCap (with $7$ classes), we trained A2M models with their default architecture.
 
 As \Cref{tab:baseline_benchmark,tab:baseline_mocap} show, we obtain significant improvements over the most competitive prior works\cite{guo2020action2motion,petrovich21actor} in all metrics, while other modified baselines lag far behind. We would like to make the following observations: (i) ACTOR's performance gain over A2M in the HumanAct12 dataset is attributed to replacing autoregressive block with Transformer in \cite{petrovich21actor}; they achieve similar results as A2M when using autoregressive blocks. We, however, use autoregressive blocks, and not only significantly outperform  A2M, but also outperform  ACTOR itself. Moreover, we achieve higher performance while using the basic $(x,y,z)$ representation, instead of more complex rotation representations. We attribute such improvements to the hierarchical decoding and action-type aware embedding in the latent space. (ii) The A2M and ACTOR models' performances are almost equal on the UESTC dataset (overlapping $95\%$ confidence intervals in Acc), while the A2M models perform slightly worse (around $2$-$3\%$) in the HumanAct12 dataset. However, for the MoCap dataset, A2M's performance is significantly worse and ODMO outperforms A2M-Lie by a clear margin of around $15\%$ in Acc. We believe that this is due to the complexity of the MoCap dataset, especially since the real motion sequences have longer spatial translations. Thus, we believe that our ODMO would show consistently high performance across different types of datasets. (iii) ODMO does far better for all of the diversity/multimodality
 metrics except the n-APD in UESTC dataset, but the A2M models have lower accuracy than ODMO's illustrating the potential artifacts generated by the A2M models in UESTC dataset. 
 
\subsubsection{Ablation Study}
\label{section:ablation}
The comprehensive ablation study should consider all of the four terms in \Cref{func: final loss}.
\textbf{Motion reconstruction loss}($-\mathcal{L}_{mre}(\psi)$): It ensures the generator regresses on motion sequences. All performance metrics are indeed very poor when this term is ablated.  \textbf{Contrastive learning} ($-\mathcal{L}_{es}(\theta)$): Without the contrastive loss in the encoder, ODMO becomes a classic VAE and its performance decreases significantly, as shown in \Cref{tab:baseline_benchmark,tab:baseline_mocap}. Thus, forming the latent space via contrastive learning is essential to ODMO. \textbf{Trajectory \& Consistency loss} ($-\mathcal{L}_{traj}$\& $-\mathcal{L}_{cons}$):  Since experiments on these two loss terms have almost no effect on the traditional metrics, we  only discuss their relevance in \Cref{tab:motion_customization,subsec:traj_cus} in the context of achieving higher customization accuracy. In addition to the loss terms, we conduct  experiments on \textbf{End point} ($-\myvec{r_e}$):
The model created by removing the end point as an input to the trajectory generator in \Cref{tab:baseline_benchmark,tab:baseline_mocap} shows comparable performance as the model with the end location (ODMO) across all three datasets. Thus,  superior performance metrics of ODMO (see  \Cref{tab:baseline_benchmark,tab:baseline_mocap}) are not primarily due to the end point feature in our decoder: even without it our architecture pipeline is able to match or exceed the performance of other baseline models. By having $\myvec{r_e}$ as an input, the ODMO provides novel customization capabilities. All ablation variants are trained using the same hyper-parameters as in ODMO.

\input{tab/mm_apd}
\subsection{Mode discovery and customization}\label{section:customization}
\subsubsection{Mode discovery and homogeneity analysis}
As explained in \Cref{sec:mm_sampling}, the ODMO's latent space enables us to discover modes within each motion category. \Cref{fig:mode_discovery} illustrates a few example modes discovered by our adaptive GMM estimation methodology from two activity categories in HumanAct12 and UESTC dataset respectively. The sampled real motion sequences from
each mode show consistent style. Once discovered, each mode can be sampled on demand to generate motion sequences of a desired type.

There is, however, a need to quantitatively measure the semantic uniformity of  real motion sequences belonging to the same modes across all categories. For any model and category $c$, let $M_c$ be the number of modes discovered for that category via GMM clustering of the latent codes of real motion sequences in the category, and let $\mathcal{G}(c,k) $ be the set of real motion sequences belonging to mode $k$ inside category $c$, and $\mathcal{G}_c = \mathcal{G}(c,1)\cup \mathcal{G}(c,2) \cdots \mathcal{G}(c, M_c)$. Then, lower the value of APD($\mathcal{G}(c,k)$) is, the more homogeneous the mode $k$ is. Then, for any mode-aware model, we expect its expected APD for category $c$: $\textrm{mode-APD}(\mathcal{G}_c) =  \frac{1}{|\mathcal{G}_c|}\sum_{k=1}^{M_c}|\mathcal{G}(c,k)|\text{APD}(\mathcal{G}(c,k))$ to be lower than  $APD(\mathcal{G}_c)$. However, for effective comparison across models one needs to compute one single average metric across all categories. Thus we define a relative metric that computes the \textit{decrease in APD} when modes are introduced:
\vspace{0.1cm}
$$\textrm{mode-homogenity}= 100 \times  \frac{1}{|C|} \sum_{c=1}^{|C|} \frac{\textrm{APD}(\mathcal{G}_c) - \textrm{mode-APD}(\mathcal{G}_{c})}{\textrm{APD}(\mathcal{G}_c)}.$$

In \Cref{tab:Homogeneity_analysis}, we compared the mode-homogeneity metric  based on the modes discovered in the latent spaces of the ODMO and ACTOR models (Action2Motion used time-varying VAE with time dependent latent codes, so we do not compare it here and MoCap results are dropped since ACTOR does not show meaningful performance in MoCap). Clearly, $\text{Modes}_\text{ODMO}$ outperforms $\text{Modes}_\text{ACTOR}$. 
Our results, thus, show that ODMO is able to construct more meaningful and structured latent space than the SOTA algorithm ACTOR.

\input{tab/traj_cus_error}
\input{figure/target_loc}
\subsubsection{
Intra-action type motion interpolation}
Once discovered,  we can interpolate between codes belonging to different mode clusters to generate new samples not present in the training dataset. As shown in \Cref{fig:mode_discovery}, we can observe the styles morphing seamlessly from one type of Warm up/dumbbell-one-arm-shoulder-pressing to another. Motion style transfer \cite{aberman2020unpaired} itself is an active research area and requires human labeling of the styles for training. Thus, besides using such intra-action type motion customization for animation purposes, our systematic data generation  can be also used to perform data augmentation for style transfer tasks.

\subsection{Trajectory customization}\label{subsec:traj_cus}
Programmable end locations are highly desirable for animation tasks and can be further combined with route planning for more complex motion generation in any given scenario. We evaluated ODMO's performance in trajectory customization  qualitatively and quantitatively. \Cref{fig:multim} visualizes a few examples where ODMO produces realistic and controllable motion sequences. For example, given a style or code, one would like the generated motion sequence to reach a specified destination in the work space scenario. Similarly, given a pair of specified points it would be desirable to generate several motion sequences that originate at one point and display different trajectories and styles while reaching the second end point. 

There are two fundamental criteria for successful completions of these tasks: generated motion needs to be realistic and reach its target successfully. At the population level, we use classifier's accuracy to measure the quality of generated motion sequences. To measure success in reaching the specified end location, we compute a goal-error metric $dist_e$ as the mean squared error between target end location $\myvec{r_e}$ and the end location in the generated motion. 
Specifically, in all three datasets, we randomly sampled $40$ codes for each motion category, and for each code we randomly sampled two real end points $\myvec{r_e}$ within that category and interpolated eight end points in between. For each sampled code, the task for our ODMO model is to generate ten motion sequences that end at the corresponding ten end locations, starting at the origin. These $400$ pairs of code and $\myvec{r_e}$ in each motion category were sent to the downstream modules to generate corresponding motion sequences. The accuracy and corresponding $dist_e$ along with the $95\%$ confidence interval in ten different trials with different random seeds for different datasets are shown in \Cref{tab:motion_customization}.
We also demonstrate the performance of ODMO variants trained with consistency loss removed ($-\mathcal{L}_{cons}$) and with both trajectory loss and consistency loss removed ($-\mathcal{L}_{traj}$).
All of the variants are able to generate comparablly realistic motion sequences considering the relatively high Acc they obtained in all datasets.
However, in term of $dist_e$, the ODMO model (with both trajectory loss and consistency loss) has the most powerful customization ability such that the generated motion can best reach the specified end locations. (See \textit{Supplementary Video} for better visualization.)

%% file: tab/mingjian_trial.tex
\begin{table*}[!ht]
\footnotesize
\centering
\caption{\label{tab:baseline_benchmark}Benchmark on HumanAct12 $\mathrm{\&}$ UESTC Dataset. ODMO obtains significant improvements
over the prior works. Note for UESTC, FID$_{tr}$ and FID$_{test}$ are computed based on train and test split as done in \cite{petrovich21actor}.}
\begin{tabular}{@{}lccccc|cccccc@{}}
\toprule
&\multicolumn{5}{c|}{HumanAct12}  & \multicolumn{6}{c}{UESTC}   \\ \midrule
Methods                     & Acc$\uparrow$        & FID$\downarrow$     & Diversity          & MModality  & n-APD  & Acc$\uparrow$        & FID$_{\textit{tr}}$ $\downarrow$ & FID$_{\textit{test}}$ $\downarrow$ & Diversity          & MModality  & n-APD        \\ \midrule
Real motions                & $\mstd{99.70}{.06}$  & $\mstd{0.02}{.01}$  & $\mstd{7.08}{.06}$ & $\mstd{2.53}{.05}$ & $\mstd{100.0}{.00}$& $\mstd{99.79}{.05}$ & $\mstd{0.01}{.00}$              &       $\mstd{0.05}{.00}$                             & $\mstd{7.20}{.06}$ & $\mstd{1.61}{.02}$ & $\mstd{100.0}{.00}$ \\ \midrule
DLow-Gen                    & $\mstd{8.29}{.16}$   & $\mstd{4.87}{.03}$  & $\mstd{6.39}{.09}$ & $\mstd{6.39}{.06}$ & $\mstd{139.6}{.68}$ & $\mstd{2.71}{.11}$  & $\mstd{3.28}{.06}$             &      $\mstd{3.01}{.06}$                             & $\mstd{6.39}{.06}$ & $\mstd{6.33}{.03}$ & $\mstd{197.1}{1.1}$\\
Act-MoCoGAN                & $\mstd{8.29}{.18}$   & $\mstd{34.8}{.06}$ & $\mstd{1.64}{.02}$ & $\mstd{1.63}{.04}$ &$\mstd{9.12}{.09}$ & $\mstd{3.87}{.07}$  & $\mstd{25.5}{.05}$             &      $\mstd{24.8}{.05}$                             & $\mstd{4.22}{.12}$ & $\mstd{2.28}{.02}$ & $\mstd{21.21}{.11}$\\
A2M-xyz                     & $\mstd{69.17}{.55}$  & $\mstd{1.73}{.06}$  & $\mstd{6.73}{.06}$ & $\mstd{4.18}{.09}$ & $\mstd{82.16}{.58}$ &$\mstd{90.25}{.21}$  & $\mstd{0.39}{.01}$ &     $\mstd{0.39}{.00}$                              & $\mstd{7.08}{.05}$ & $\mstd{1.83}{.02}$ &$\mstd{\bf{79.97}}{.43}$\\
A2M-Lie                     & $\mstd{93.25}{.47}$  & $\mstd{0.21}{.01}$  & $\mstd{7.02}{.05}$ & $\mstd{2.77}{.07}$ & $\mstd{85.34}{.61}$ &$\mstd{91.07}{.21}$  & $\mstd{0.24}{.01}$              &        $\mstd{0.25}{.00}$                           & $\mstd{7.09}{.06}$ & $\mstd{1.91}{.04}$ & $\mstd{79.54}{.39}$ \\
ACTOR                       & $\mstd{95.86}{.28}$  & $\mstd{0.13}{.00}$  & $\mstd{7.04}{.06}$ & $\mstd{2.46}{.07}$ & $\mstd{74.30}{.43}$& $\mstd{91.82}{.18}$  & $\mstd{0.23}{.01}$              &      $\mstd{0.26}{.01}$                             & $\mstd{7.10}{.09}$ & $\mstd{1.89}{.04}$ &$\mstd{74.10}{.33}$ \\ \midrule
\rowcolor[HTML]{FFF7E6}
ODMO                        & $\mstd{\bf{97.81}}{.21}$  & $\mstd{\bf{0.12}}{.01}$  & $\mstd{\bf{7.05}}{.15}$ & $\mstd{\bf{2.57}}{.04}$ & $\mstd{\bf{96.88}}{.72}$ &$\mstd{\bf{93.67}}{.18}$ & $\mstd{\bf{0.15}}{.02}$ &$\mstd{\bf{0.17}}{.00}$ 
&$\mstd{\bf{7.11}}{.06}$ & $\mstd{\bf{1.61}}{.02}$ & $\mstd{75.38}{.18}$ \\ 
\midrule
ODMO$_{conv}$ & $\mstd{98.08}{.15}$  & $\mstd{0.11}{.05}$  & $\mstd{7.02}{.13}$ & $\mstd{2.58}{.05}$ & $\mstd{97.30}{.94}$&
$\mstd{92.86}{.43}$ & $\mstd{0.18}{.01}$ &
$\mstd{0.21}{.01}$&
$\mstd{7.10}{.06}$ & $\mstd{1.63}{.02}$ & $\mstd{70.22}{.45}$ \\ 
$-\myvec{r_e}$          & $\mstd{97.07}{.19}$  & $\mstd{0.12}{.01}$  & $\mstd{7.03}{.08}$ & $\mstd{2.56}{.07}$ & $\mstd{86.84}{.79}$ &$\mstd{93.13}{.01}$  & $\mstd{0.14}{.04}$&    $\mstd{0.16}{.00}$ & $\mstd{7.16}{.06}$ & $\mstd{1.74}{.04}$ &  $\mstd{69.52}{.28}$ \\
$-\mathcal{L}_{es}(\theta)$ & $\mstd{75.07}{1.5}$ & $\mstd{0.82}{.11}$  & $\mstd{6.84}{.06}$ & $\mstd{4.12}{.10}$ &$\mstd{83.68}{3.1}$& $\mstd{84.04}{.45}$ & $\mstd{0.38}{.03}$ &        $\mstd{0.35}{.03}$& $\mstd{6.97}{.10}$ & $\mstd{2.51}{.07}$ & $\mstd{65.21}{1.1}$ \\
\bottomrule
\end{tabular}
\end{table*}

%% file: tab/mocap.tex
\begin{table}[]
\footnotesize
\centering
\caption{\label{tab:baseline_mocap}Benchmark on MoCap Dataset. ODMO obtains significant improvements over
prior works in all metrics.}
\begin{tabular}{@{}lccccc@{}}
\toprule
                            & \multicolumn{4}{c}{CMU MoCap}                                                        \\ \midrule
Methods                     & Acc$\uparrow$        & FID$\downarrow$     & Diversity          & MModality  & n-APD         \\ \midrule
Real motions                & $\mstd{98.54}{.14}$  & $\mstd{0.02}{.00}$  & $\mstd{6.57}{.11}$ & $\mstd{2.17}{.08}$ & $\mstd{100.0}{.00}$ \\ \midrule
DLow-Gen                    & $\mstd{14.4}{.17}$  & $\mstd{10.9}{.05}$ & $\mstd{5.13}{.08}$ & $\mstd{5.06}{.17}$  & $\mstd{249.4}{3.1}$ \\
Act-MoCoGAN                 & $\mstd{14.93}{.09}$  & $\mstd{37.2}{.12}$ & $\mstd{0.90}{.04}$ & $\mstd{0.84}{.04}$ &$\mstd{8.41}{.18}$ \\
A2M-xyz                     & $\mstd{76.17}{.67}$  & $\mstd{1.32}{.05}$  & $\mstd{6.33}{.07}$ & $\mstd{3.27}{.08}$ & $\mstd{77.49}{1.1}$ \\
A2M-Lie                     & $\mstd{76.33}{.66}$  & $\mstd{1.42}{.06}$  & $\mstd{6.28}{.08}$ & $\mstd{3.35}{.07}$ & $\mstd{72.83}{1.3}$\\
ACTOR & $\mstd{15.21}{1.2}$  & $\mstd{2.68}{.23}$  & $\mstd{6.20}{.11}$& $\mstd{6.02}{.14}$ &  $\mstd{124.5}{3.4}$\\\midrule
\rowcolor[HTML]{FFF7E6} 
ODMO                        & $\mstd{\bf{93.51}}{.04}$  & $\mstd{\bf{0.34}}{.03}$  & $\mstd{\bf{6.56}}{.07}$ & $\mstd{\bf{2.49}}{.06}$ & $\mstd{\bf{88.13}}{1.2}$\\
\midrule
ODMO$_{conv}$ & $\mstd{92.99}{.28}$  & $\mstd{0.38}{.03}$  & $\mstd{6.54}{.07}$& $\mstd{2.44}{.08}$ &   $\mstd{90.40}{1.5}$\\
$-\myvec{r_e}$          & $\mstd{86.58}{.39}$  & $\mstd{0.99}{.07}$  & $\mstd{6.31}{.08}$ & $\mstd{2.89}{.12}$ &$\mstd{72.82}{.99}$\\
$-\mathcal{L}_{es}(\theta)$ & $\mstd{69.90}{2.8}$ & $\mstd{2.01}{.24}$  & $\mstd{6.10}{.11}$ & $\mstd{3.65}{.14}$& $\mstd{67.89}{3.8}$ \\
\bottomrule
\end{tabular}
\end{table}

%% file: figure/demo_mm_fig.tex
\begin{figure*}[!h]
    \centering
    \includegraphics[width = 0.85\textwidth]{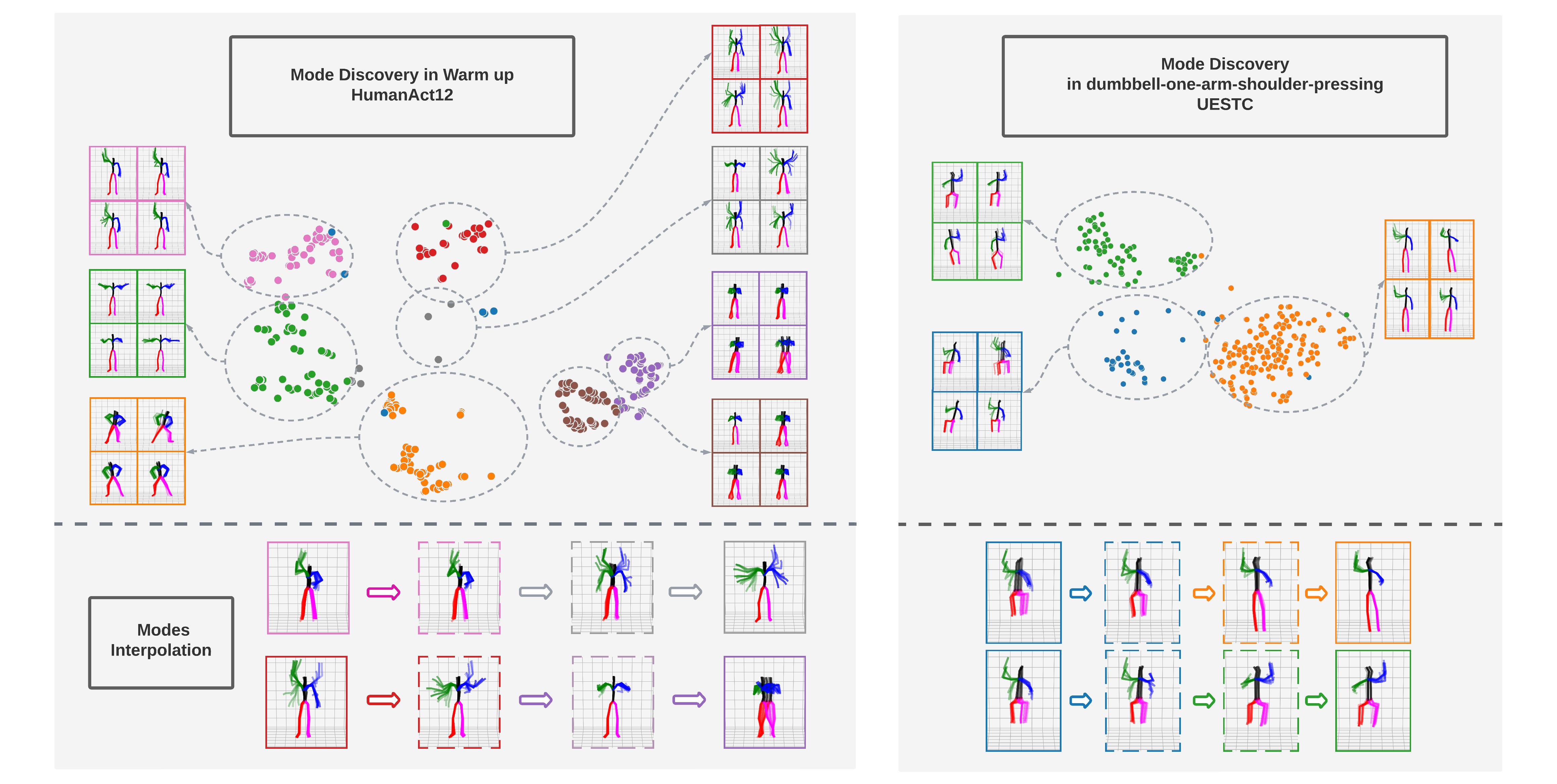}
    \caption{\textbf{Mode discovery in Warm Up (HumanAct12, Left) and in  dumbbell-one-arm-shoulder-pressing  (UESTC, Right), and Intra-Action Type Motion Interpolations}: 
    Latent codes are projected into a 2D space using t-SNE \cite{van2008visualizing} to illustrate that the codes can be separated into distinct clusters (the discovered modes from GMM are color coded).  For each cluster, we sample a few codes and display their corresponding motion sequences via time-lapsed images. These time-lapsed images are highly consistent within clusters, providing evidence that each cluster represents a distinct mode of the action type. As the bottom insets demonstrate, interpolations between codes from two modes have smooth morphing. 
    }
    \label{fig:mode_discovery}
\end{figure*}

%% file: tab/mm_apd.tex
\begin{table}[]
\centering
\footnotesize
\caption{\label{tab:Homogeneity_analysis} Homogeneity analysis of the latent space; higher mode-homogenity means the modes discovered are purer. The higher mode-homogenity of ODMO indicates that it constructs more meaningful latent space than ACTOR.
}
\begin{tabular}{@{}c|c|c@{}}
\toprule                     & HumanAct12                & UESTC \\ \midrule
Method       & \multicolumn{1}{c|}{mode-homogenity$\uparrow$  } & mode-homogenity $\uparrow$ \\ \midrule
Real Motions  &$\mstd{0.00\%}{.00}$ &$\mstd{0.00\%}{.00}$\\
$\text{Modes}_{\text{ACTOR}}$&$\mstd{0.14\%}{.07}$&$\mstd{0.05\%}{.02}$\\
\rowcolor[HTML]{FFF7E6} 
$\text{Modes}_{\text{ODMO}}$  &$\mstd{\bf{15.22}\%}{.92}$&$\mstd{\bf{8.74}\%}{.57}$      \\ \bottomrule
\end{tabular}
\end{table}

%% file: tab/traj_cus_error.tex
\begin{table}[]
\centering
\footnotesize
\caption{\label{tab:motion_customization} Evaluation of trajectory customization at the population level shows that ODMO  generates realistic motion (as judged by Acc) and reaches close to the desired end location (as measured by $dist_e$). The ablation study of $-\mathcal{L}_{cons}$ and $-\mathcal{L}_{traj}$ demonstrates the effectiveness of the combination of trajectory loss in \Cref{eq:traj_gen} and consistency loss in \Cref{eq:cons} for trajectory customization.}
\begin{tabular}{@{}lcc|cc|cc@{}}
\toprule
& \multicolumn{2}{c|}{Mocap} & \multicolumn{2}{c|}{HumanAct12} & \multicolumn{2}{c}{UESTC} \\ \midrule
Method      & Acc $\uparrow$          & $dist_e$   $\downarrow$        & Acc  $\uparrow$            & $dist_e$    $\downarrow$          & Acc  $\uparrow$         & $dist_e$    $\downarrow$       \\ \midrule
$-\mathcal{L}_{traj}$  & $\mstd{89.23}{.59}$ &$\mstd{0.71}{.04}$& $\mstd{96.02}{.35}$ &$\mstd{0.04}{.00}$&  $\mstd{92.84}{.32}$ &$\mstd{0.03}{.01}$ \cr 
$-\mathcal{L}_{cons}$ & $\mstd{88.04
}{.41}$   &$\mstd{0.35}{.03}$& $\mstd{94.21}{.34}$ &$\mstd{0.02}{.00}$& $\mstd{\bf{94.02}}{.24}$  &$\mstd{0.02}{.01}$\cr
\rowcolor[HTML]{FFF7E6} 
ODMO  & $\mstd{\bf{91.56}}{.40}$ &$\mstd{\bf{0.06}}{.00}$& $\mstd{\bf{96.95}}{.23}$  &$\mstd{\bf{0.00}}{.00}$&$\mstd{93.42}{.22}$  &$\mstd{\bf{0.01}}{.00}$\cr\bottomrule
\end{tabular}
\end{table}

%% file: figure/target_loc.tex
\begin{figure*}[!h]
    \centering
      \includegraphics[width =0.49\textwidth]{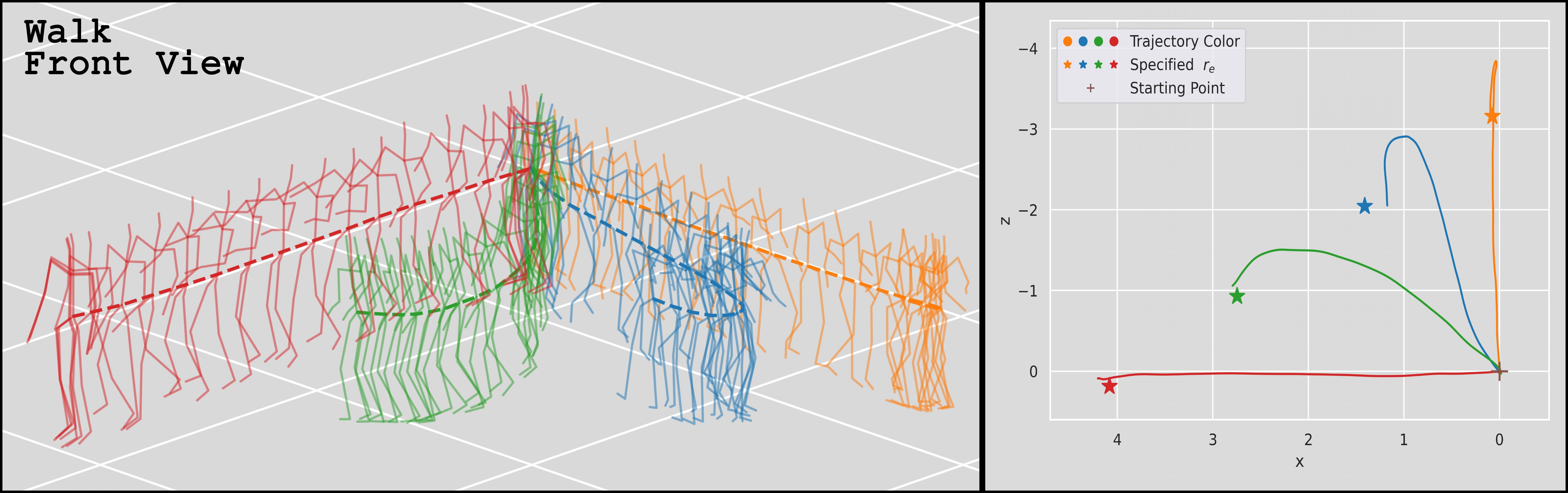}
    \includegraphics[width =0.49\textwidth]{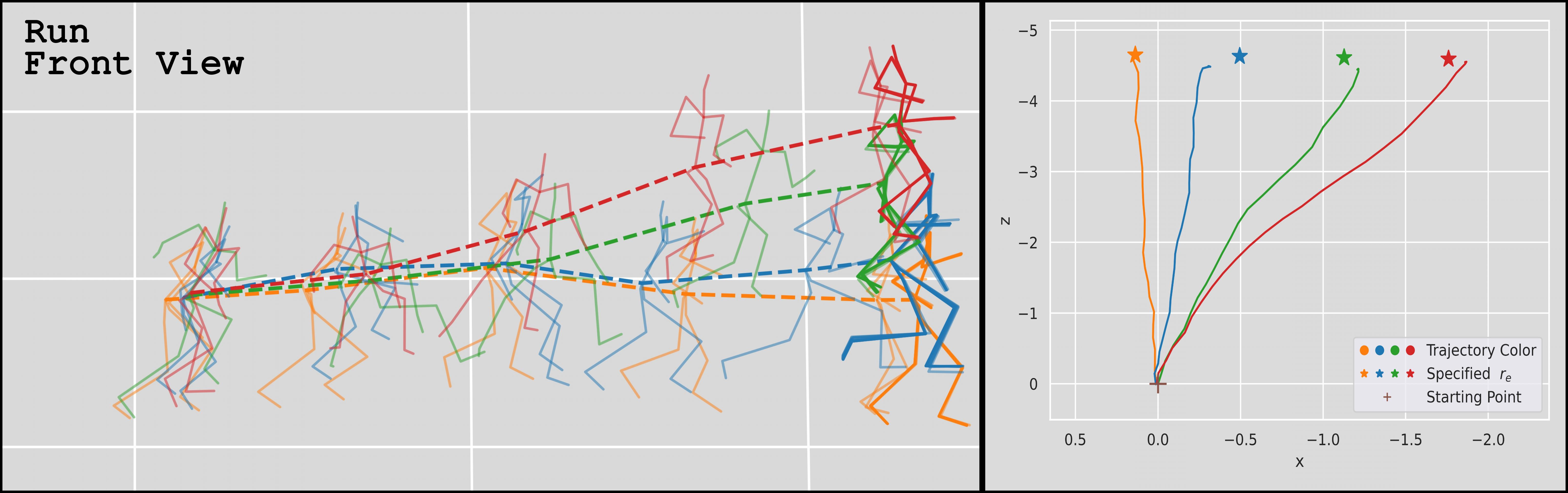}
    \includegraphics[width =0.49\textwidth]{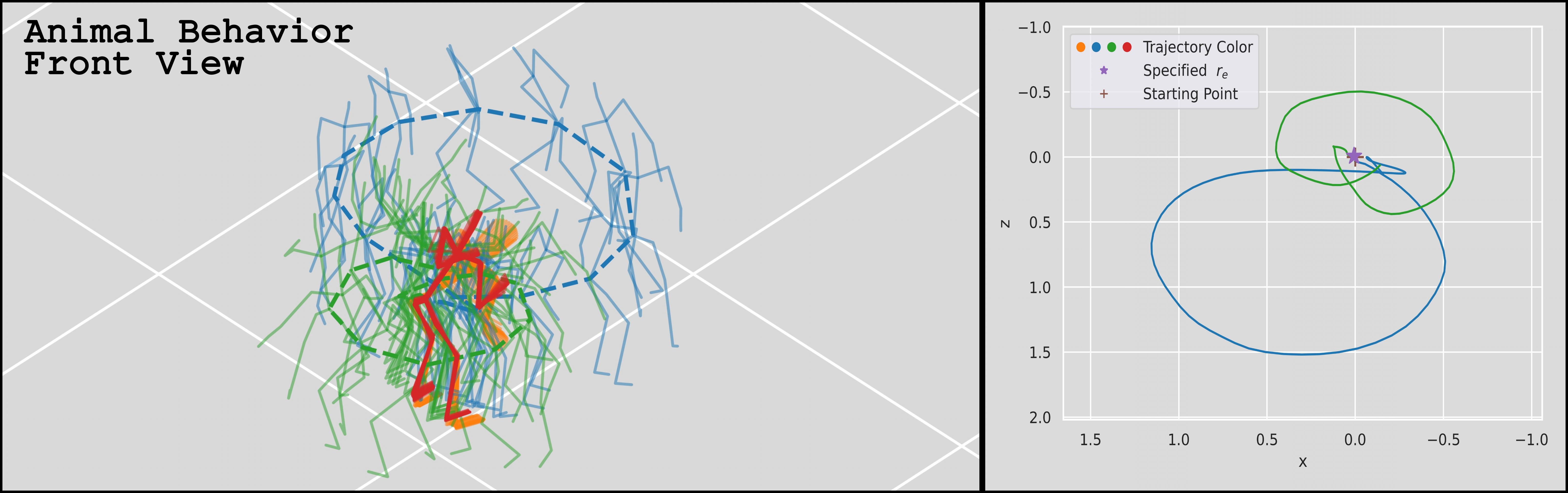}
    \includegraphics[width =0.49\textwidth]{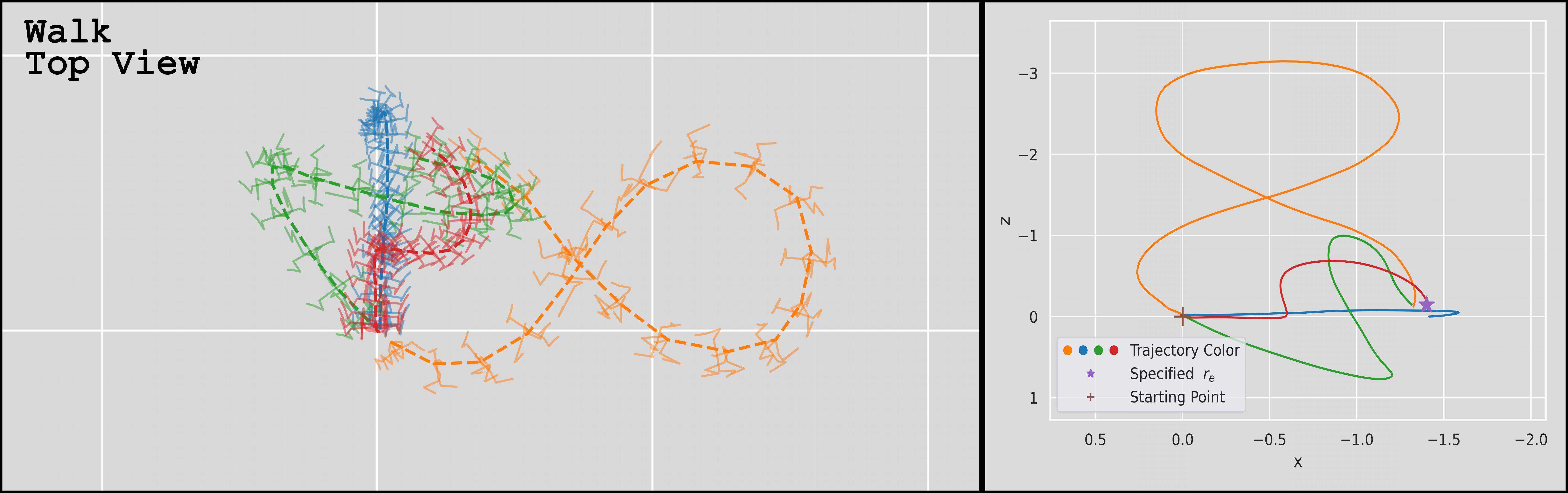}    
    \caption{Examples of on-demand realistic motion generation with end point specified trajectory customization. Examples are taken from different categories in MoCap dataset and
    the viewpoints are rotated for better visualization. Vertical view of motion trajectory and the specified end points are also provided for reference. Top two examples: two unseen end points are created by linear interpolation of two end points from real data. Then a randomly sampled code in the given category is paired with each end location for motion sequence generation. Bottom two examples: in each example, one end point is sampled. Then each of four sampled codes from the corresponding action category is paired with the end point for motion sequence generation. These demonstrate ODMO’s capability to generate multiple motion sequences for a specified end location. 
    \label{fig:multim}}
\end{figure*}

%% file: sec/5_conclusions.tex
\section{Concluding Remarks}
We propose a novel framework, On-Demand MOtion Generation (ODMO), for generating realistic, diverse and customised long-term 3D human motion sequences conditioned only on action type.
While recent work \cite{guo2020action2motion,petrovich21actor} has focused on this task, customization capabilities such as the following have not been investigated: (i) ability to automatically discover different modes/styles of motion within each action type in the training set, and then generate samples from these modes on demand; (ii) ability to interpolate modes within an action type to generate diverse motions not seen in the training set, and (iii) ability to customize trajectories and destination locations of generated motion sequences. Such new capabilities would widen the spectrum of potential applications for 3D human motion generators. \textit{Extensive examples of customization are included in the Supplementary Materials}.

%% file: sec/supplimentary.tex
\newcommand{\beginsupplement}{%
        \setcounter{table}{0}
        \renewcommand{\thetable}{S\arabic{table}}%
        \setcounter{figure}{0}
        \renewcommand{\thefigure}{S\arabic{figure}}%
        \setcounter{equation}{0}
        \setcounter{page}{1}
        \setcounter{section}{0}
     }

\pagebreak

\begin{center}
\LARGE{\textbf{Supplementary Material for: \\Action-conditioned On-demand \\ Motion Generation}}
\end{center}

\beginsupplement 
\section{Overview}
Our supplementary material includes two parts:
\begin{itemize}
    \item Supplementary video which includes gif examples for mode interpolation and trajectory customization.
    \item Supplementary document which introduces more implementations details as mentioned in the main text.
\end{itemize}

In this document, we first introduce the model architectures and more implementation details in   \Cref{sec:im_details}, then demonstrate more details for our mode preserving VAE in \Cref{sec:mode_VAE}. Dataset details can also be found in \Cref{sec:dataset}. 

\section{Implementation Details}\label{sec:im_details}
In this section, we introduce the implementation details of ODMO and other baseline algorithms, mode-preserving sampling, and metrics for evaluating the generated motion sequences. 
\subsection{ODMO Implementation Detail}
We implemented the framework of ODMO using PyTorch and trained the models with TITAN RTX GPU. 
The detailed architecture for each component is shown in Figure 2 in main text and \Cref{arch:odmo}. Note that the latent space dimension (the output of the LMP encoder) is $20$ and the decoder only relied on this small code space and the one-hot encoded class information to reconstruct the motion. During training, the contrastive margin $\alpha$ is set to be $5$, and the target variance of posteriors is set to $0.05$. 
In the training of the network, we adopted the Adam optimizer with the learning rate $10^{-3}$ for both dataset. Additionally, teacher forcing (TF) is applied for enforcing the stable and efficiency of the training with the TF rate linear decayed from $1$ to $0.3$.
We also provide the option of rendering the generated skeleton with SMPL-X model, as shown in Figure 1 in the main text. As a post-processing step, the output skeleton sequences go through a optimization module to generate the body meshes which align with SMPL-X model.
\subsection{Mode-preserving sampling and silhouette score}
We use sihouette score to find the number of modes in activity category adaptively. To be specific, we fit a GMM with $k$ mixture components and compute the silhouette score $S_k$. We vary $k$ from $3$ to $11$ to get $S_3 \cdots S_{11}$. Starting from $S_3$, we determine an $i$ such that $S_i > 0.95*S_{i+1}$ for the first time. Then the number of modes is selected as $i$. We dropped mixture components whose size is smaller than $10$ for denoising purpose. 

\subsection{Finding the optimal $\sigma^2$}
As we show in \Cref{sec:mode_VAE}, contrastive mechanism applied on the encoder's output mean $\mu(\textbf{x})$ together with the prior regularization of encoder's output covariance vector $\sigma^2(\textbf{x})$ helps to form our data-driven mode-aware latent space. To find the most suitable range of $\sigma^2$ (the hyper-parameter in $\sigma^2(\textbf{x})$'s prior  distribution) for the motion generation task, we sweep the parameter and list the result in \Cref{tab:mm_sigma}. As we can see, as $\sigma^2$ increases, the performance drops especially for Acc and FID since the estimated posterior $q_{\phi}(\mathbf{z\mid x})$ becomes too flat losing essential information of each motion $\mathbf{x}$. At the same time, when $\sigma^2$ becomes too small, the generalization ability of the learnt model decreases thus Diversity is low in general. Taking all the metrics different models obtained and the effect of $\sigma^2$ on model generalization ability into consideration, we use $\sigma^2=0.05$ to train all our final models (listed in the main paper). 
\input{sm_tab/select_sigma}
\input{sm_tab/model_arch}
\input{sm_tab/classifier_arch}

\subsection{Baseline Algorithms}

\textbf{Action2Motion (A2M):}
We use A2M as one of our major baselines for the three public datasets. We trained the A2M with joints' xyz coordinates representation (A2M xyz) and A2M using Lie group representation (A2M Lie) on MoCap dataset for seven motion categories with their default parameters. On HumanAct12 dataset, we directly employed their released model as baseline. On UESTC dataset, we also directly trained A2M xyz and A2M Lie with their default parameters. 

\textbf{ACTOR:} We treat ACTOR as the other strong baselines for UESTC and HumanAct12 dataset. We used their released model for final evaluation. 

\textbf{Other baselines:}
For implementation of Act-MoCoGAN, we followed their default architecture while replacing the image generator with pose generator and modified the image, video discriminators accordingly, as introduced in A2M paper. We trained the Act-MoCoGAN by tweaking the hyper-parameters so that it achieves the best generative results in the three dataset.  As for the Gen-DLow, we adapted it to motion generation task by replacing the motion history condition as the class one-hot encoding. We trained the Gen-DLow (both VAE and DLow) in much more epochs (5000,600) than the default parameter (500, 500) so that it can be better to achieve the generative task.    

\subsection{Metrics Computation Details}
Our metric computation follows the same procedure as Action2-Motion (A2M) which includes sampling of ground truth and generated motion sequences, as well as computing fidelity and diversity (multimodality) metrics based on them. However, since A2M and ACTOR used slightly different sampling strategies on generating the final evaluation metrics, for fair comparison, we recomputed their metrics with our sampling setup instead of using the numbers reported in their papers. 
\subsubsection{Metrics based on classifier}
Frechet Inception Distance(FID), Multimodality and Diversity are computed based on the extracted features, i.e the output of the feature layer (see \Cref{arch:mocap_classifier,arch:uestc_classifier}) in the classifier.
\textbf{FID}: We randomly sample (400/250/200) samples for (MoCap / HumanAct12 / UESTC) samples from ground truth motion and generated motion across all categories indepedently. Then we calculate the sample mean $\mu$ and covariance matrix $\Sigma$ of the activation of the extracted features  $x \in \mathbb{R}^{30}$ (the feature dimension in classifier according to Action2Motion) from the set of real motion sequences as $(\mu_1, \Sigma_1)$ and the set of generated motion sequences as $(\mu_2, \Sigma_2)$ respectively. Lastly, FID is computed as $\norm{\mu_1 - \mu_2}^2 + \Tr(\Sigma_1 + \Sigma_2 - 2\sqrt{\Sigma_1  \Sigma_2})$.

\textbf{Multimodality}: To measure diversity inside one category, we randomly sample $20$ pairs of motion sequences inside that category and extract their features as set $\mathcal{C}_1$, then compute the average pairwise Euclidean distance among these pairs, i.e, $\frac{1}{20}\sum_{(x_i, x_j)\in \mathcal{C}_1} \norm{x_i - x_j}$.

\textbf{Diversity}: To measure the diversity of motion sequences across motion categories, we randomly sample $200$ pairs of motion across all categories and extract their features as set $\mathcal{C}_2$, then calculate the average pairwise Euclidean distance, i.e.$\frac{1}{200}\sum_{(x_i, x_j)\in \mathcal{C}_2} \norm{x_i - x_j}$.

\vspace{-0.1cm}
\subsubsection{Motion Classifiers}
The metrics based on motion classifiers trained on real motion have been introduced in the A2M paper and we applied the same framework to compare performance of different models.  For HumanAct12 dataset, we used the released classifier from A2M to make the results we benchmarked comparable. For MoCap dataset, we used 7 action categories instead of 8 action categories as in the A2M paper (Wash activity is dropped due to the noisy data), and thus we need to train our own classifier for our MoCap dataset. The architecture is shown in \Cref{arch:mocap_classifier}. Similarly, for UESTC dataset, a xyz representaion based classifier is not avalable from previous works, so we trained our own classifier with architecture shown in \Cref{arch:uestc_classifier}. 

\section{Mode preserving VAE}\label{sec:mode_VAE}
\label{sec:contrastive_mechanism}
\subsection{VAE: From variational perspective to prior regularization}
From a Variational Bayesian perspective, the evidence low bound (ELBO) can be derived from data log likelihood $\log p_\theta(x)$ minus KL divergence between the estimated posterior $q_{\phi}(\mathbf{z} \mid \mathbf{x})$ and $p_{\theta}(\mathbf{z} \mid \mathbf{x})$ as shown below:
\begin{align*}
\log p_{\theta}(\mathbf{x}) &=\mathbb{E}_{\mathbf{z} \sim q_{\phi}(\mathbf{z} \mid \mathbf{x})}\left[\log p_{\theta}(\mathbf{x})\right] \\
&=\mathbb{E}_{\mathbf{z} \sim q_{\phi}(\mathbf{z} \mid \mathbf{x})}\left[\log \left(\frac{p_{\theta}(\mathbf{x}, \mathbf{z})}{q_{\phi}(\mathbf{z} \mid \mathbf{x})} \cdot \frac{q_{\phi}(\mathbf{z} \mid \mathbf{x})}{p_{\theta}(\mathbf{z} \mid \mathbf{x})}\right)\right] \nonumber \\
&=\underbrace{\mathbb{E}_{\mathbf{z} \sim q_{\phi}(\mathbf{z} \mid \mathbf{x})}\left[\log \left(\frac{p_{\theta}(\mathbf{x}, \mathbf{z})}{q_{\phi}(\mathbf{z} \mid \mathbf{x})}\right)\right]}_{ELBO}+ D_{KL}(q_{\phi}(\mathbf{z} \mid \mathbf{x})||p_{\theta}(\mathbf{z} \mid \mathbf{x})).
\end{align*}

To train a VAE for a given dataset $D$, the loss function, which needs to be minimized, is the negative of the ELBO, and can be rewritten as 
\begin{align*}
-l_{\theta, \phi}(D) &= \sum_{\mathbf{x}\in D} \mathbb{E}_{\mathbf{z} \sim q_{\phi}(\mathbf{z} \mid \mathbf{x})}\left[\log \left(p_{\theta}(\mathbf{x}, \mathbf{z})\right)  -\log \left(q_{\phi}(\mathbf{z} \mid \mathbf{x})\right)\right] \nonumber\\
& = \sum_{\mathbf{x}\in D} \mathbb{E}_{\mathbf{z} \sim q_{\phi}(\mathbf{z} \mid \mathbf{x})}[\log(p_{\theta }(\mathbf {x\mid z} ))]+D_{KL}(q_{\phi }(\mathbf {z\mid x} )\parallel p_{\theta }(\mathbf {z} )).
\end{align*}

For the standard VAE, we assume the prior $p(\mathbf{z})$ is a standard multivariate Gaussian distribution with zero mean and unit variance, and the estimated posterior $q_\phi(\mathbf{z\mid x})$ is assumed to be a normal distribution with  mean $\mathbf{\mu}(\mathbf{x}) \in \mathbb{R}^d$ and covariance $\Sigma(\mathbf{x}) = \mathrm{diag}(\sigma^2(\mathbf{x}))\in \mathbb{R}^{d\times d}$. Taking the powerful transformation capacity of neural networks into consideration, this assumption generally works. With this choice of $q_\phi(\mathbf{z\mid x})$,  we have:

\begin{align}
    -l_{\theta, \phi}(D) &=\sum_{\mathbf{x}\in D} \mathbb{E}_{\mathbf{z} \sim q_{\phi}(\mathbf{z} \mid \mathbf{x})}[\log(p_{\theta }(\mathbf {x\mid z} ))]+D_{KL}(q_{\phi }(\mathbf {z\mid x} )\parallel p_{\theta }(\mathbf {z} ))\nonumber\\
    & = \sum_{\mathbf{x}\in D}( \mathbb{E}_{\mathbf{z} \sim q_{\phi}(\mathbf{z} \mid \mathbf{x})}[\log(p_{\theta }(\mathbf {x\mid z} ))]+ \nonumber\\
    &\hspace{0.3cm}\frac{1}{2}\left[-\sum_{i}^d\left(\log \sigma_{i}^{2}(\mathbf{x})+1\right)+\sum_{i}^d \sigma_{i}^{2}(\mathbf{x})+\sum_{i}^d \mu_{i}^{2}(\mathbf{x})\right]).
    \label{eq:vae_loss-3}
\end{align}

The encoder and decoder networks with parameters, $\phi$ and $\theta$, respectively can be trained jointly with the loss function in \Cref{eq:vae_loss-3}. In summary, this loss function includes two parts: (i) reconstruction loss; (ii) KLD between $\mathcal{N}(0, I)$ and $\mathcal{N}(\mu(\mathbf{x}), \Sigma(\mathbf{x}))$ which in turn decomposes into two terms (a) 
$\mu(\mathbf{x})^T\mu(\mathbf{x})$ , and (b) a term involving $\Sigma(\mathbf{x}))$: $-\ln(\det \Sigma(\mathbf{x})) + \Tr(\Sigma(\mathbf{x}))$.

If we leave the variational framework aside, ELBO can also be viewed from a quite different optimization perspective: VAE is more like an autoencoder with additional sampling process in the latent space. We want the encoding decoding process have good reconstruction results, so a reconstruction loss term should be imposed. At the same time, we want some regularization on the encoder's output $\mu(\mathbf{x})$ and $\sigma(\mathbf{x})$ such that there is less overfitting and more generalization. Without regularization of $\mu(\mathbf{x})$, the latent space can potentially be infinite and too sparse and specific for each datapoint. Without regularization of $\sigma(\mathbf{x})$, it can safely go to 0 which reduces to an autoencoder without any sampling power. So we need regularization for both terms. Thus we can rewrite ELBO as a combination of a reconstruction loss and a regularization term on $\mu(\mathbf{x})$ and $\sigma(\mathbf{x})$, i.e. the negative of log of prior distribution of these parameters.

We assume that for each data point $\mathbf{x}$, the prior for encoded vector $\mu$ is independent over all components. In the standard VAE, for $\mu$, that term would correspond to a prior of $\mu$ under normal distribution, i.e. $$\frac{-1}{2} \mu_{i}^2(\mathbf{x}) \rightarrow \mu_{i}(\mathbf{x}) \sim \mathcal{N}\left(\mu_{i}(\mathbf{x}) \mid 0,1\right).$$ Similarly, for $\sigma(\mathbf{x})$, its related regularization term would correspond to a prior of gamma distribution with a mean of 3:
$$
\frac{1}{2}\left(\log \left(\sigma_{i}^2(\mathbf{x})\right)-\sigma_{i}^2(\mathbf{x})\right) \rightarrow \sigma_{i}^2(\mathbf{x}) \sim \Gamma\left(\sigma_{i}^2(\mathbf{x}) \mid \alpha=\frac{3}{2}, \beta=\frac{1}{2}\right).
$$

Starting from this observation, we modify the priors for both $\mu(\mathbf{x})$ and $\sigma(\mathbf{x})$ in our contrastive VAE framework for more expressive latent space. 
For the $\sigma_i^2(x)$, we scale the hyper-parameter of the Gamma distribution shown below so that it has a lower mean. This is represented by the hyperparameter $\sigma^2$ in Equation (1) in the main text. 

$$
\frac{1}{2}\left(\log \left(\sigma_{i}^2(\mathbf{x})\right)-\frac{1}{\sigma^2}\sigma_{i}^2(\mathbf{x})\right) \rightarrow \sigma_{i}^2(\mathbf{x}) \sim \Gamma\left(\sigma_{i}^2(\mathbf{x}) \mid \alpha=\frac{3}{2}, \beta=\frac{1}{2\sigma^2}\right).$$

For the $\mu(\mathbf{x})$ , instead of having independent Gaussian priors for different data points, we use the contrastive mechanism to generate a joint prior that clusters the s belonging to the same action category. This leads to our hierarchical mode-aware latent space. Its details can be found in \Cref{sec:contrastive_mech}

\subsection{Contrastive Mechanism}\label{sec:contrastive_mech}

Following this point of view, more complex prior of $\mu$'s should be constructed to provide more modeling capacity in the latent space. To automatically form such a structure in the latent space, we deducted some general principles for the algorithm to follow. First, when the motion sequences are similar, their latent codes should be nearby, it leads to the assumptions that the latent codes corresponding to the same mode should cluster in the latent space.
At the same time, the low-dimensional representation of different motion sequences are far apart in the latent space, which means that each category should its own cluster.

Instead of using standard distributions, we form a distribution using self contrasting mechanism. We want to regularize $\mu$ jointly for $n$ datapoints. The most convenient way is to view it as a spring ensemble system. The probability of one configuration $X:(\mu_1,\cdots, \mu_n)$ is modeled as Boltzmann distribution, which is proportional to the exponential of the configurations energy, i.e. $p(x) \propto e^{-\beta E}$, where $E=\sum_{i=1}^n\sum_{j=1}^n \frac{1}{2}k(c_i, c_j)(\mu(\mu_i, \mu_j) - \mu_s(c_i, c_j))^2$. In different cases, we specify stiffness function $k$ and stress free length $\mu_s$ accordingly. 

For encoded two data points corresponding to the same activity, $\mu_i$ and $\mu_j$, then we assign a spring whose stress free length is $\mu_s=0$ and the dynamic length is $\mu =\norm{\mu_i - \mu_j}$. We assume the stiffness is 1, then its energy is $\frac{1}{2}(\mu - \mu_s)^2$. 
Higher energy is contained by the pair of points with further distance. A spring like this is assigned to every pair of $\mu_i$ and $\mu_j$ belonging to the same category for all categories. Thus if we look at energy from one category $c$, minimizing this part of the energy would be equivalent to minimizing the distance of each $\mu_{c,i} (1 \leq i \leq n_c)$ (encoded data belonging to category $c$) to the category center $\mu_c = \frac{\sum_{i=1}^{n_{c}} \mu_{c,i}}{n_c}$, as shown in \Cref{eq: gm_center}.

\begin{align}
\sum_{i=1}^{n_c}\sum_{j=1}^{n_c} \frac{1}{2}\norm{\mu_{c,i} - \mu_{c,j}}^2 = 2n_c \sum_{i=1}^{n_c}  \frac{1}{2} \norm{\mu_{c,i} - \mu_c}^2
\label{eq: gm_center}
\end{align}

Considering the encoded two data points $\mu_i$  and $\mu_j$ corresponding to different activities, we define a compression-only spring with stress free length $\mu_s = \alpha$ (margin size), and the spring's dynamic length as $\mu= \norm{\mu_i - \mu_j}$, similarly as in the previous case. For this compression only spring, the energy is $\frac{1}{2}(\mu - \mu_s)^2$ only if $\mu < \alpha$. This leads to a hinge loss function in the optimization. 

\section{Datasets} \label{sec:dataset}
\input{sm_fig/skeleton}
We used CMU MoCap dataset, HumanAct 12 dataset, and UESTC dataset, which contain diversified 3D motion sequences with varied lengths for seven, twelve, and forty daily activities respectively. MoCap dataset represents 3D human skeleton as 21 joints and 20 bones; while HumanAct 12 dataset uses 24 joints and 23 bones; For UESTC dataset, there are 18 joints and 17 bones as shown in \Cref{fig:human_ske}. The root node for these three datasets is $0$. The detailed categories and number of motions belonging to each dataset are shown in \Cref{tab:mocap_data,tab:HA_data,tab:UESTC_data}. 

During the training phase, in order to balance the training samples from different motion categories, we applied the stratified sampling strategy.  Furthermore, since each real motion sequence in the training data has different time duration, in each training epoch, we randomly sample a consecutive motion segment with the desired length from each real motion sequence: $100$ frames for MoCap and $60$ frames for HumanAct12, and $60$ frames for UESTC.  If one motion sequence is shorter than the desired length, we pad it with the last frame until the desired length is reached. After obtaining the fixed-length sub-sequences from the original data source, we shift it to ensure that the initial frame's root joint is at the origin.

\input{sm_tab/dataset}
\input{sm_tab/dataset_HA}
\input{sm_tab/dataset_uestc}

%% file: sm_tab/select_sigma.tex
\begin{table}[!ht]
\footnotesize
\centering
\caption{\label{tab:mm_sigma} $\sigma^2$ selection in three datasets.}
\begin{tabular}{@{}lcccc@{}}
\toprule
&\multicolumn{4}{c}{HumanAct12}   \\ \midrule
Methods                     & Acc$\uparrow$        & FID$\downarrow$     & Diversity          & MModality                   \\ \midrule
Real motions                & $\mstd{99.70}{.06}$  & $\mstd{0.02}{.01}$  & $\mstd{7.08}{.06}$ & $\mstd{2.53}{.05}$  \\ \midrule
$\sigma^2 = 6.25$                        & $\mstd{94.06}{.30}$  & $\mstd{0.30}{.01}$  & $\mstd{6.98}{.15}$ & $\mstd{2.57}{.05}$  \\
$\sigma^2 = 1.25$                        & $\mstd{95.84}{.36}$  & $\mstd{0.31}{.01}$  & $\mstd{7.00}{.15}$ & $\mstd{2.53}{.07}$  \\
$\sigma^2 = 0.25$                        & $\mstd{97.52}{.22}$  & $\mstd{0.15}{.01}$  & $\mstd{7.08}{.14}$ & $\mstd{2.42}{.05}$  \\
$\sigma^2 = 0.05$                        & $\mstd{97.81}{.21}$  & $\mstd{0.12}{.01}$  & $\mstd{7.05}{.15}$ & $\mstd{2.57}{.04}$  \\
$\sigma^2 = 0.01$                        & $\mstd{97.81}{.21}$  & $\mstd{0.09}{.02}$  & $\mstd{7.03}{.13}$ & $\mstd{2.53}{.09}$  \\
\bottomrule
\end{tabular}

\footnotesize
\centering
\begin{tabular}{@{}lcccc@{}}
\toprule
&\multicolumn{4}{c}{Mocap}   \\ \midrule
Methods                     & Acc$\uparrow$        & FID$\downarrow$     & Diversity          & MModality                   \\ \midrule
Real motions                & $\mstd{98.54}{.14}$  & $\mstd{0.02}{.00}$  & $\mstd{6.57}{.11}$ & $\mstd{2.17}{.08}$ \\ \midrule
$\sigma^2 = 6.25$                        & $\mstd{86.95}{.56}$  & $\mstd{1.14}{.05}$  & $\mstd{6.32}{.09}$ & $\mstd{2.85}{.10}$  \\
$\sigma^2 = 1.25$                        & $\mstd{89.05}{.37}$  & $\mstd{0.89}{.05}$  & $\mstd{6.40}{.08}$ & $\mstd{2.63}{.08}$  \\
$\sigma^2 = 0.25$                        & $\mstd{91.69}{.38}$  & $\mstd{0.60}{.02}$  & $\mstd{6.51}{.07}$ & $\mstd{2.63}{.07}$  \\
$\sigma^2 = 0.05$                        & $\mstd{93.51}{.39}$  & $\mstd{0.34}{.03}$  & $\mstd{6.56}{.07}$ & $\mstd{2.49}{.06}$  \\
$\sigma^2 = 0.01$                        & $\mstd{89.82}{.26}$  & $\mstd{0.80}{.03}$  & $\mstd{6.40}{.07}$ & $\mstd{2.83}{.08}$  \\
\bottomrule
\end{tabular}
\tiny
\begin{tabular}{@{}lccccc@{}}
\toprule
&\multicolumn{5}{c}{UESTC}   \\ \midrule
Methods                     & Acc$\uparrow$        & FID$_{tr}$ $\downarrow$ & FID$_{test}$$\downarrow$    & Diversity          & MModality                   \\ \midrule
Real motions                & $\mstd{99.79}{.05}$  & $\mstd{0.01}{.00}$              &       $\mstd{0.05}{.00}$                             & $\mstd{7.20}{.06}$ & $\mstd{1.61}{.02}$  \\ \midrule
$\sigma^2 = 6.25$                        & $\mstd{59.19}{.22}$  & $\mstd{3.46}{.03}$  & $\mstd{3.45}{.03}$ & $\mstd{6.51}{.07}$ & $\mstd{2.26}{.03}$  \\
$\sigma^2 = 1.25$                     & $\mstd{77.18}{.28}$  & $\mstd{1.25}{.02}$  & $\mstd{1.27}{.02}$ & $\mstd{6.80}{.04}$ & $\mstd{2.09}{.03}$  \\
$\sigma^2 = 0.25$                     & $\mstd{89.83}{.20}$  & $\mstd{0.44}{.01}$  & $\mstd{0.40}{.01}$ & $\mstd{6.99}{.07}$ & $\mstd{1.87}{.03}$  \\
$\sigma^2 = 0.05$                        & $\mstd{93.67}{.18}$  & $\mstd{0.15}{.00}$  & $\mstd{0.17}{.00}$ & $\mstd{7.11}{.07}$ & $\mstd{1.61}{.03}$  \\
$\sigma^2 = 0.01$                     & $\mstd{93.70}{.17}$  & $\mstd{0.10}{.00}$  & $\mstd{0.12}{.00}$ & $\mstd{7.08}{.06}$ & $\mstd{1.76}{.05}$  \\
\bottomrule
\end{tabular}
\end{table}

%% file: sm_tab/model_arch.tex
\begin{table*}
\footnotesize
\center
\caption{\label{arch:odmo}Architecture of ODMO}
\begin{tabular}{@{}l|l|l@{}}
\toprule
Module                                & Layers          & details                                 \\ \midrule
LMP encoder        & RNN layer       & 1 LSTM layer with 64 hidden units \\ \cline{2-3} 
                                      & Feature layer   & 2 fully connected layers followed by BN and PReLU activation \\ \cline{2-3} 
                                      & Output layer    & 1 fully connected layer                                      \\ \midrule
Trajectory generator & Embedding layer & 2 fully conencted layers with PReLU activation in between    \\ \cline{2-3} 
                                      & RNN layer       & 2 LSTM layers with 128 hidden units                          \\ \cline{2-3} 
                                      & Feature layer   & 3 fully conencted layers with PReLU activation               \\ \cline{2-3} 
                                      & Output layer    & 1 fully connected layer                                      \\ \midrule
Motion generator                      & Embedding layer & 1 fully conencted layer with PReLU activation                \\ \cline{2-3} 
(Trajectory encoder)                  & RNN layer       & 2 LSTM layers with 128 hidden units                          \\ \midrule
Motion generator                      & Embedding layer & 1 fully conencted layer with PReLU activation                \\ \cline{2-3} 
(Motion decoder)                      & RNN layer       & 2 LSTM layers with 128 hidden units                          \\ \cline{2-3} 
                                      & Output layer    & 1 fully connected layer                                      \\ \bottomrule
\end{tabular}
\end{table*}

%% file: sm_tab/classifier_arch.tex
\begin{table}[]\footnotesize
\centering
\caption{\label{arch:mocap_classifier}Architecture of MoCap and HumanAct12 motion classifiers (same as Action2Motion)}
\begin{tabular}{@{}l|l@{}}
\toprule
Layers        & Details                                    \\ \midrule
RNN layer     & 2 GRU layers with 128 hidden units         \\\midrule
Feature layer & fully connected layer with Tanh activation \\\midrule
Output layer  & 1 fully connected layer                    \\ \bottomrule
\end{tabular}
\end{table}

\begin{table}[]
\footnotesize
\centering
\caption{\label{arch:uestc_classifier}Architecture of our UESTC classifier}
\begin{tabular}{@{}l|l@{}}
\toprule
Layers                           & Details                                         \\ \midrule
RNN layer                      & 3 GRU layers with 256 hidden units                         \\ \midrule
Feature layer  & fully connected layer with Tanh activation    \\ \midrule
Output layer                   & 1 fully connected layer                         \\ \bottomrule
\end{tabular}
\end{table}

%% file: sm_fig/skeleton.tex
\begin{figure}[!h]
    \centering
    \includegraphics[width=0.5\textwidth]{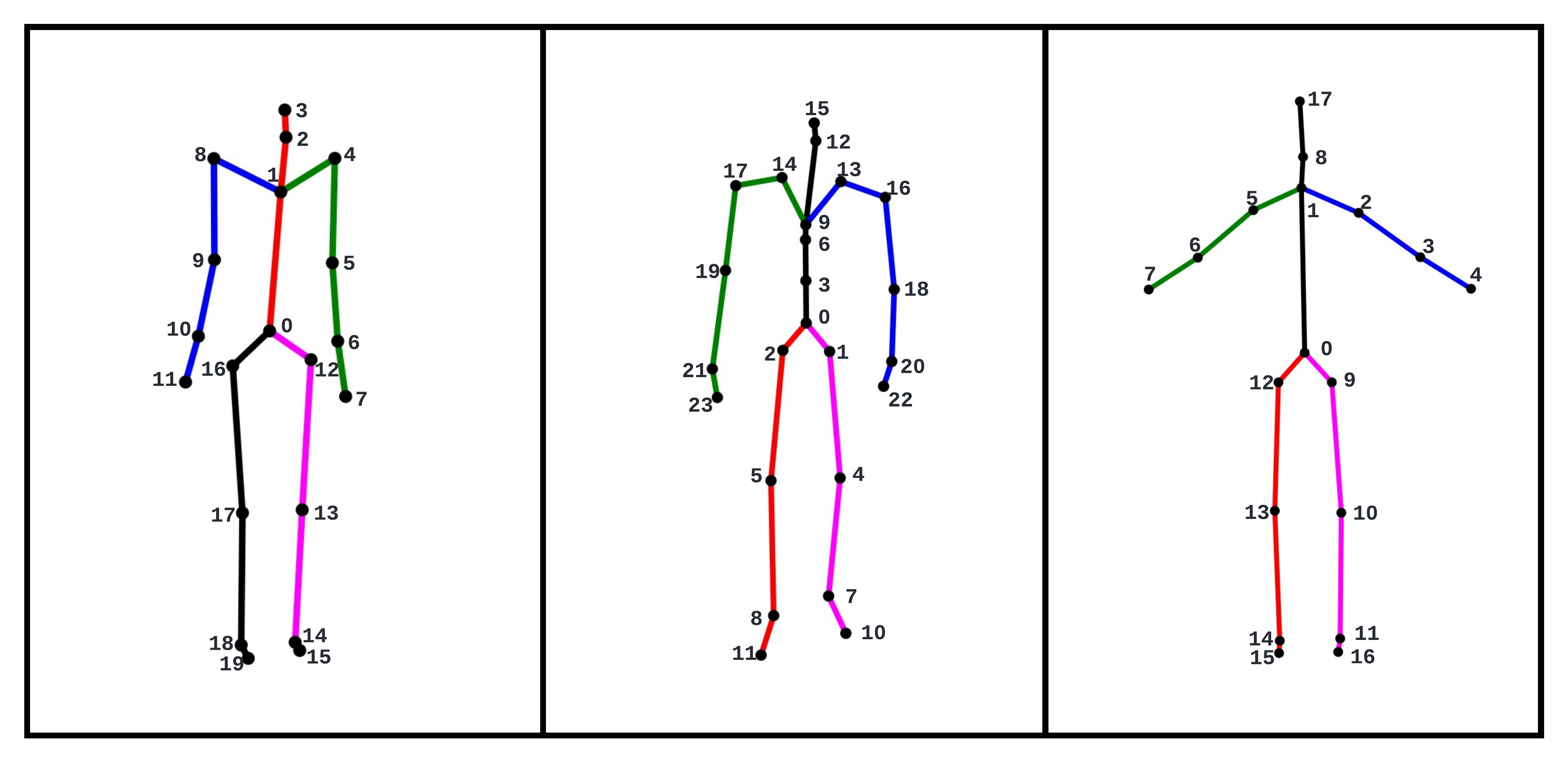}
    \caption{Human skeleton in the MoCap dataset (left), HumanAct12 dataset (middle), and UESTC dataset (right) with the joint number annotated in black. Root node (pelvis) is annotated as 0.}
    \label{fig:human_ske}
\end{figure}

%% file: sm_tab/dataset.tex
\begin{table}[!b]
\footnotesize
\centering
\caption{Statistics of our MoCap Dataset\label{tab:mocap_data}}
\begin{tabular}{@{}|cc|}
\hline
Action type & Number of sequences \\
\hline
Walk & 474 \\
Run & 109 \\
Dance & 77\\
Jump & 108\\
Animal Behavior & 101\\
Step & 68\\
Climb & 33\\
\hline
Total & 970\\
\hline
\end{tabular}
\end{table}

%% file: sm_tab/dataset_HA.tex
\begin{table}[!b]
\centering
\footnotesize
\caption{Statistics of HumanAct12 Dataset \label{tab:HA_data}}
\begin{tabular}{@{}|cc|}
\hline
Action type & Number of sequences \\
\hline
Warm up & 215 \\
Walk & 47 \\
Run & 50\\
Jump & 94\\
Drink & 88\\
Lift dumbbell & 218\\
Sit & 54\\
Eat & 77\\
Turn steering wheel & 56\\
Phone & 61\\
Boxing & 140\\
Throw & 91\\
\hline
Total & 1191\\
\hline
\end{tabular}
\end{table}

%% file: sm_tab/dataset_uestc.tex
\begin{table}[!hb]
\footnotesize
\centering
\caption{Statistics of UESTC Dataset \label{tab:UESTC_data}}
\begin{tabular}{@{}|cc|}
\hline
Action type & Number of sequences \\
\hline
standing-gastrocnemius-calf            &          345 \\
single-leg-lateral-hopping             &          345 \\
high-knees-running                     &          345 \\
rope-skipping                          &          343 \\
standing-toe-touches                   &          339 \\
front-raising                          &          285 \\
straight-forward-flexion               &          285 \\
standing-opposite-elbow-to-knee-crunch &          285 \\
dumbbell-side-bend                     &          285 \\
shoulder-raising                       &          285 \\
single-dumbbell-raising                &          285 \\
wrist-circling                         &          285 \\
punching                               &          285 \\
pulling-chest-expanders                &          284 \\
shoulder-abduction                     &          281 \\
overhead-stretching                    &          270 \\
head-anticlockwise-circling            &          270 \\
deltoid-muscle-stretching              &          270 \\
upper-back-stretching                  &          270 \\
spinal-stretching                      &          268 \\
alternate-knee-lifting                 &          255 \\
knee-to-chest                          &          255 \\
knee-circling                          &          255 \\
bent-over-twist                        &          255 \\
standing-rotation                      &          255 \\
pinching-back                          &          240 \\
dumbbell-shrugging                     &          240 \\
dumbbell-one-arm-shoulder-pressing     &          240 \\
elbow-circling                         &          240 \\
arm-circling                           &          235 \\
raising-hand-and-jumping               &          225 \\
forward-lunging                        &          225 \\
left-kicking                           &          225 \\
left-lunging                           &          225 \\
punching-and-knee-lifting              &          225 \\
squatting                              &          225 \\
jumping-jack                           &          225 \\
marking-time-and-knee-lifting          &          225 \\
rotation-clapping                      &          225 \\
left-stretching                        &          224 \\
\hline
Total & 10629\\
\bottomrule
\end{tabular}
\end{table}